\documentclass[journal,comsoc]{IEEEtran}
\usepackage[T1]{fontenc}
\ifCLASSINFOpdf
\else
\fi
\usepackage[cmintegrals]{newtxmath}
\usepackage{amsmath}
\usepackage[ruled]{algorithm2e}
\usepackage{times}
\usepackage{epsfig}
\usepackage{graphicx}
\usepackage{subfigure}
\usepackage{amsmath}

\usepackage{amssymb}
\usepackage{multirow}
\usepackage{verbatim}
\usepackage{bm}
\usepackage{framed}
\usepackage{makecell}
\usepackage{booktabs}
\usepackage{multirow}
\usepackage{url}

\begin{document}	
	
	\title{SFace: Sigmoid-constrained Hypersphere Loss \\ for Robust Face Recognition}
	\author{Yaoyao Zhong,
		Weihong Deng, 
		Jiani Hu,
		Dongyue Zhao,
		Xian Li,
		Dongchao Wen
		\thanks{Yaoyao Zhong, Weihong Deng, and Jiani Hu are with the Pattern Recognition and Intelligent System
			Laboratory, School of Artificial Intelligence, Beijing University of Posts and Telecommunications, Beijing 100876, China (e-mail: zhongyaoyao@bupt.edu.cn; whdeng@bupt.edu.cn; jnhu@bupt.edu.cn). Weihong Deng is the corresponding author. }
		\thanks{Dongyue Zhao, Xian Li, and Dongchao Wen are with Canon Information Technology (Beijing) Co., Ltd. (e-mail: zhaodongyue@canon-ib.com.cn; lixian@canon-ib.com.cn; wendongchao@canon-ib.com.cn).}
	}
	\maketitle
	
\begin{abstract}
Deep face recognition has achieved great success due to large-scale training databases and rapidly developing loss functions. The existing algorithms devote to realizing an ideal idea: minimizing the intra-class distance and maximizing the inter-class distance. However, they may neglect that there are also low quality training images which should not be optimized in this strict way. Considering the imperfection of training databases, we propose that intra-class and inter-class objectives can be optimized in a moderate way to mitigate overfitting problem, and further propose a novel loss function, named sigmoid-constrained hypersphere loss (SFace). Specifically, SFace imposes intra-class and inter-class constraints on a hypersphere manifold, which are controlled by two sigmoid gradient re-scale functions respectively. The sigmoid curves precisely re-scale the intra-class and inter-class gradients so that training samples can be optimized to some degree. Therefore, SFace can make a better balance between decreasing the intra-class distances for clean examples and preventing overfitting to the label noise, and contributes more robust deep face recognition models. Extensive experiments of models trained on CASIA-WebFace, VGGFace2, and MS-Celeb-1M databases, and evaluated on several face recognition benchmarks, such as LFW, MegaFace and IJB-C databases, have demonstrated the superiority of SFace.    
\end{abstract}
\IEEEpeerreviewmaketitle

\section{Introduction}

\begin{figure*}[htbp]
	\center
	\includegraphics[width=0.99\linewidth]{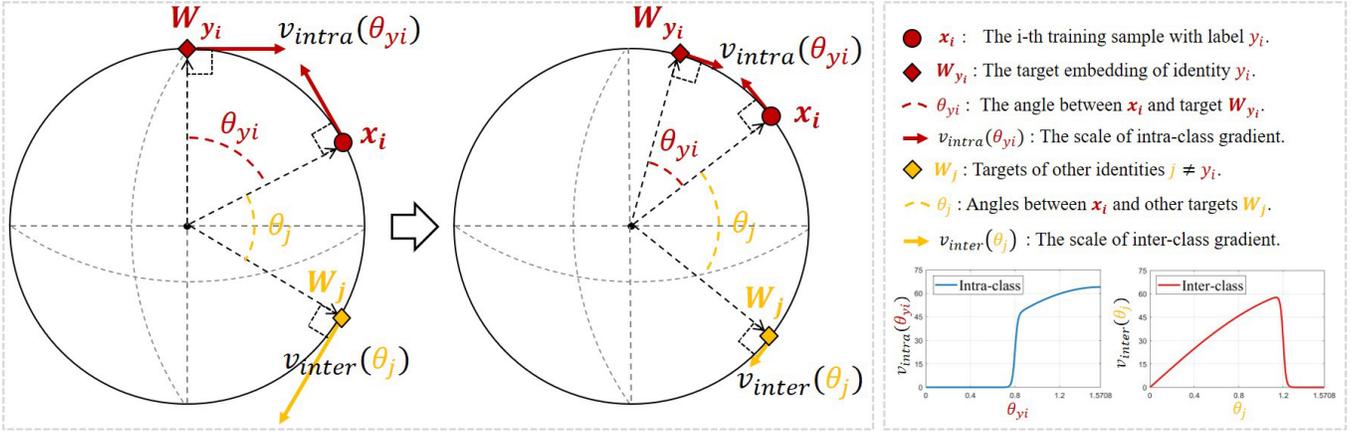}
	\caption{Schematic illustration of the sigmoid-constrained hypersphere loss, which imposes intra-class and inter-class constraints on a hypersphere manifold. The optimizing directions of samples and target embedding are always along the tangent of the hypersphere while the moving speed is controlled by two sigmoid curves respectively. Specifically, the moving speed of the deep feature $\bm{{x_i}}$ and its target center ${\bm{W_{{y_i}}}}$ decreases gradually as they approaching to each other, while the moving speed of $\bm{{x_i}}$ and other target centers ${\bm{W_{{j}}}}$ increases rapidly as they start approaching to each other.}
	\label{fig:intro_illu}
\end{figure*}

\IEEEPARstart{D}{eep} face recognition has obtained surprising improvement recent years~\cite{Sun2014Deep,Schroff2015FaceNet,Wen2016A,wang2017normface,Liu2017SphereFace,Wang2018CosFace,deng2019arcface,zhang2019adacos,zhang2019p2sgrad}. The pipeline for deep face recognition has been widely used for its practical usage~\cite{zhou2015naive,wang2017normface,Liu2017SphereFace,zhang2019adacos}. That is, deep face recognition models are trained on web-collected databases~\cite{Yi2014CASIA,guo2016msceleb,nech2017level,Cao18,wang2018devil}, and work as deep feature extractors to evaluate on other testing databases~\cite{LFWTech,Wolf2011Face,zheng2017CALFW,CPLFWTech,kemelmacher2016megaface,klare2015pushing,maze2018iarpa}. 

The large-scale training databases~\cite{Yi2014CASIA,guo2016msceleb,nech2017level,Cao18,wang2018devil} are fundamental for the success of deep face recognition. For training databases of deep face recognition, we can never expect to obtain a ``perfect'' training database which should include, but not limited to, sufficient numbers of identities, and adequate images of each identity. Considering the copyright and privacy protection, the number of identities in the web-collected training databases is limited compared with the global population, and celebrities of web-collected databases may be far from the testing settings in daily life~\cite{zhou2015naive}. In addition, we can hardly collect images with full intra-class variation to model the large pose, face expressions and illumination variance of each identity~\cite{deng2017fine,maze2018iarpa}, therefore there are a significant portion of under-represented identities~\cite{Zhang2017rangeloss,yin2019feature,liu2019adaptiveface,zhong2019unequal}. Considering the open-set protocal and the limitations of training databases, current research focus is trying to make best use of the training databases, and improve the ability of loss functions to obtain a more discriminative feature extractor. One of the most effective loss functions is the large margin loss function~\cite{Liu2017SphereFace,Wang2018CosFace,deng2019arcface,zhang2019adacos,zhang2019p2sgrad}. They incorporate large margins to softmax loss to encourage the intra-class compactness and the inter-class orthogonality, which has alleviated the aforementioned quantity limitation and imbalance problem of identities to some degree.

Existing mainstream methods devote to minimizing the intra-class distance and maximizing the inter-class distance. Despite the success, they may neglect that, in addition to the high quality training images, there are also low quality training images such as misaligned images, low-resolution images, and label noise, which cannot provide effective information for distinguishing the labeled identity. Even human annotations are not reliable as we thought, because humans often struggle to distinguish between hard examples and low quality training images, and they have already been surpassed by deep face recognition models a few years ago~\cite{lu2015surpassing}. For this reason, although training databases have been elaborated by semi-automatic data cleaning algorithms~\cite{Yi2014CASIA,Cao18, wang2018devil,deng2019arcface}, there still exists noise inevitably. Due to the imperfection of training databases, strictly minimizing the intra-class distance and maximizing the inter-class distance would lead to overfitting. Therefore, our aim is to design a new loss function, which can increase the possibility of finding the best compromise between underfitting and overfitting to a specific training database, in order to obtaining better generalization ability.

Considering the imperfection of the training databases, formally, we abandon the softmax-based loss while start from the primary and fundamental idea: optimize intra-class and inter-class distances to some extent, to improve the generalization ability of models. Furthermore, we propose a novel loss function, named sigmoid-constrained hypersphere loss (SFace), to implement this idea. SFace imposes intra-class and inter-class constraints on a hypersphere manifold. The intra-class and inter-class constraints are controlled by two sigmoid curves. The sigmoid curves precisely re-scale intra-class and inter-class gradients so that intra-class and inter-class distances are optimized to some extent. As illustrated in Figure~\ref{fig:intro_illu}, for the deep feature $\bm{{x_i}}$ of a training sample, the optimizing direction is always along the tangent of the hypersphere while the moving speed is controlled by the designed gradients precisely. Specifically, the moving speed of the deep feature $\bm{{x_i}}$ and its target center ${\bm{W_{{y_i}}}}$ decreases gradually as they approaching to each other, while the moving speed of $\bm{{x_i}}$ and other target centers ${\bm{W_{{j}}}}$ increases rapidly as they start approaching to each other.

\begin{figure}[htbp]
	\centering
	\subfigure[]{
		\begin{minipage}[b]{0.17\textwidth}
			\includegraphics[width=1\textwidth]{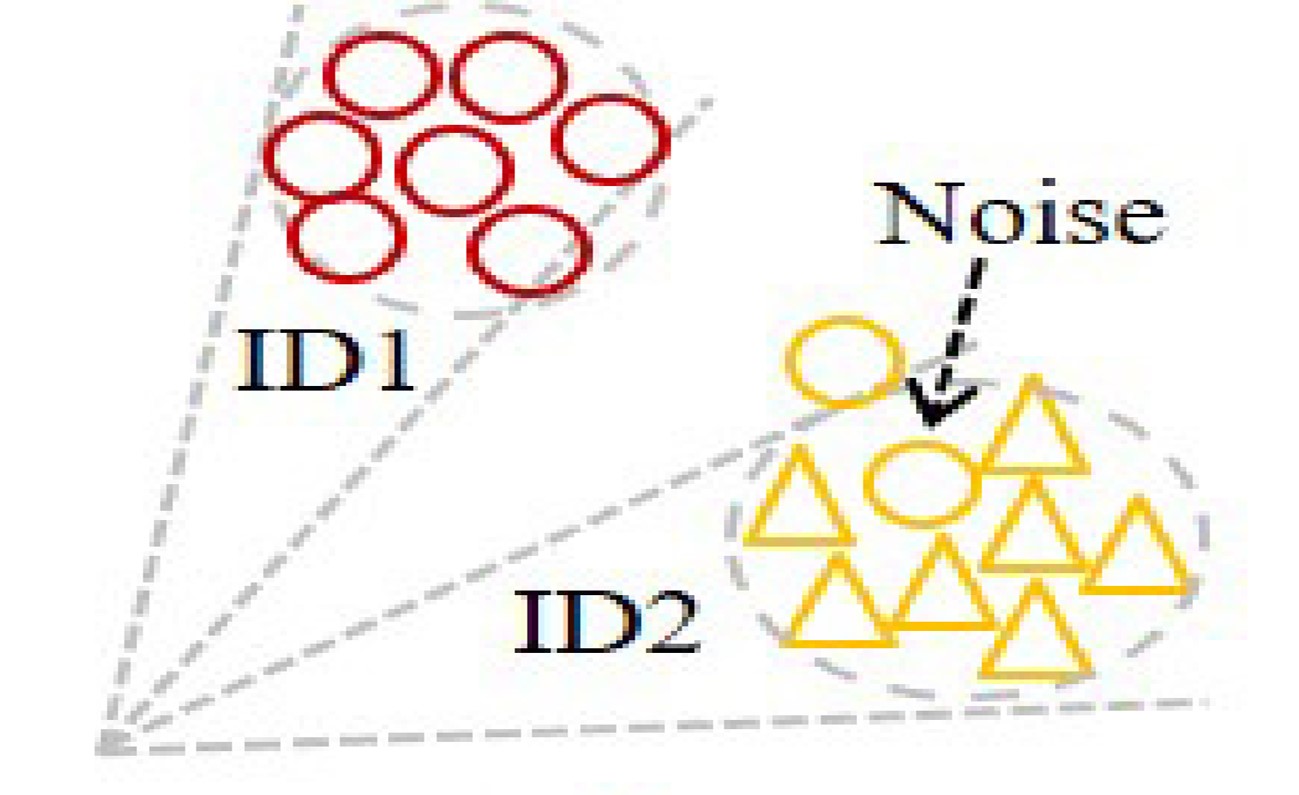}\\
	\end{minipage}}\subfigure[]{
		\begin{minipage}[b]{0.17\textwidth}
			\includegraphics[width=1\textwidth]{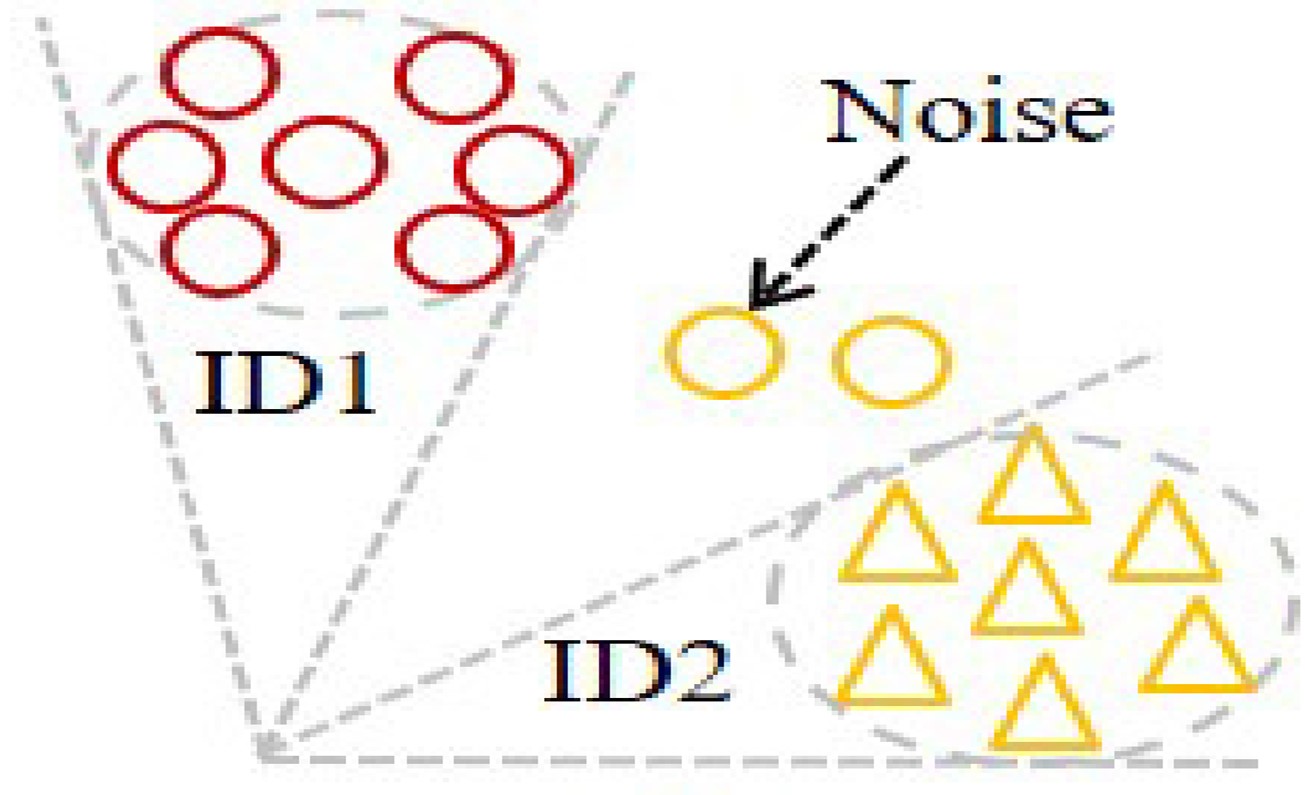}\\
			
		\end{minipage}
	}
	\caption{(a) The model would overfit to the label noise by strictly dragging the noisy samples to the wrong labeled identities. (b) In contrast, SFace can mitigate this problem in some degree because it optimizes samples in a moderate way.}
	\label{fig:intro_noise}
\end{figure}

Compared with optimizing training samples strictly, the advantage of SFace is that it provides a relatively better balance between overfitting and underfitting, for the reason that SFace adopts sigmoid functions of intra-class and inter-class gradient re-scale terms to achieve excellent control respectively. We give a simple and easy example in Figure~\ref{fig:intro_noise} for understanding. Under the label noise setting, the model would overfit to the label noise by strictly dragging the noisy samples to the wrong labeled identities. In contrast, SFace can mitigate this problem in some degree because it optimizes noisy samples in a moderate way. With the precisely control, the clean training samples are optimized earlier and more easily, while the label noise can be left behind. 

Our major contributions can be summarized as follows:
\begin{itemize}
\item{Considering the imperfection of face training databases, we introduce a new idea: optimizing intra-class and inter-class objectives in a moderate way to mitigate overfitting problem to face training databases.}
\item{Under the guidance of this idea, we propose a new loss function, named sigmoid-constrained hypersphere loss (SFace), which can increase the possibility of finding the best compromise between underfitting and overfitting, in order to obtaining better generalization ability.}
\item{Our method is evaluated on three training databases including CASIA-WebFace~\cite{Yi2014CASIA}, VGGFace2~\cite{Cao18} and MS-Celeb-1M~\cite{guo2016msceleb}, and consistently outperforms the state-of-the-art methods on several benchmarks including LFW~\cite{LFWTech}, YTF~\cite{Wolf2011Face}, CALFW~\cite{zheng2017CALFW}, CPLFW~\cite{CPLFWTech}, MegaFace~\cite{kemelmacher2016megaface}, IJB-A~\cite{klare2015pushing} and IJB-C~\cite{maze2018iarpa} databases. }
\end{itemize}

The remainder of the paper is organized as follows. Section~\ref{sec:related} briefly reviews the related deep face recognition works. In Section~\ref{sec:method}, we first give a general introduction to the proposed sigmoid-constrained hypersphere loss (SFace). Then, we detail the gradient re-scale function of SFace. Finally, we discuss the relationship between SFace and softmax based loss functions. Experimental settings and results are presented in Section~\ref{sec:exp}. Section~\ref{sec:conclusion} summarizes the conclusions.

\section{Related Work}
\label{sec:related}
In this section, we discuss and compare the loss functions in deep face recognition, which are almost entirely around the idea of minimizing the intra-class distance and maximizing the inter-class distance. There are mainly two types. 

The first type applies metric learning method in deep learning~\cite{Sun2014Deep,Schroff2015FaceNet,Wen2016A}, which maps face images to a deep feature space and directly optimizes distances, so that the inter-class distance is larger than the intra-class distance. The contrastive loss~\cite{Sun2014Deep}, triplet loss~\cite{Schroff2015FaceNet} and N-pair loss~\cite{sohn2016improved} are early methods to enhance the discrimination ability of deep features, which optimize intra-class and inter-class variance by using face pairs. Combined with softmax loss, centerloss~\cite{Wen2016A} obtains promising performance by simultaneously learns a center for deep features of each class and minimizes the distances between training samples and their corresponding
class centers. Then, range loss~\cite{Zhang2017rangeloss} minimizes overall intra-personal differences and maximizes inter-personal differences in one mini-batch. Marginal loss~\cite{Deng2017Marginal} is further proposed to maximize the inter-class distance and minimize the intra-class distance simultaneously by focusing on the marginal samples. 

The second type makes modification on cross-entropy loss (usually referred to as ``softmax loss'') to learn more discriminative features~\cite{wang2017normface,Liu2017SphereFace,deng2019arcface}. Some early works incorporate weights or features normalization~\cite{Ranjan17,wang2017normface,Zheng18ring}. L2-softmax~\cite{Ranjan17} is proposed to add an L2-constraint to the deep features and restrict them to lie on a hypersphere of a fixed radius. NSoftmax~\cite{wang2017normface} is proposed to normalize both features and weights of the last inner-product layer. Ring loss~\cite{Zheng18ring} applies soft normalization by gradually learning to constrain the norm to the scaled unit circle while preserving convexity. Then, based on previous works~\cite{Ranjan17,wang2017normface}, the large margin~\cite{Liu2016Large,Wang2018CosFace,deng2019arcface} is introduced to obtain better discriminative power by further enforcing the extra intra-class compactness and inter-class discrepancy simultaneously. L-Softmax~\cite{Liu2016Large} first incorporates a large margin to softmax loss to learn discriminative face features by strictly separating the hard samples. Instead of the multiplicative margin, CosFace~\cite{Wang2018CosFace} and ArcFace~\cite{deng2019arcface} introduce the additive margin to guarantee the convergence, which is easy for implementation. However, AdaCos~\cite{zhang2019adacos} and P2SGrad~\cite{zhang2019p2sgrad} point that the inflexible form of softmax based loss functions lacks the ability to precisely supervise the cosine distances, and they improve the large margin angular loss functions by setting the direct mapping relation between classification probability and cosine distances, which can further decrease the intra-class angles of training databases. MV-Softmax~\cite{wang2019mis} is proposed to improve softmax based loss functions by mining the mis-classified samples and emphasizing them to guide the discriminative feature learning. CurricularFace~\cite{huang2020curricularface} further develops MV-Softmax by incorporating curriculum learning, which automatically emphasizes easy samples first and hard samples later. Recent works~\cite{deng2019arcface,he2019softmax} also point that inter-class and intra-class objectives of softmax based loss functions would interact and lead to relaxation on each other. Although recent works have pointed out some shortcomings of softmax based loss functions, overall, weight/feature normalization softmax-based loss functions and large margin softmax based loss functions have significantly boosted the performance of deep face recognition. 

Our method can be categorized as the first type method in the form of metric learning, which directly optimizes the intra-class and inter-class distances. However, it also has a close connection to the second type based on softmax loss, which we will discuss in details in Section~\ref{subsection:relation}. In addition, there are also some works~\cite{Hu_2019_CVPR,Wang_2019_ICCV} aiming to solve the noise-robust training in deep face recognition, which usually use training databases with high-level label noise to obtain comparable performance with the model trained with clean databases. While our work is devoted to improving performance of models trained on clean databases which have been refined by semi-automatic data cleaning algorithms~\cite{Yi2014CASIA,Cao18,deng2019arcface}.

\section{Methodology}
\label{sec:method}
\subsection{Sigmoid-constrained Hypersphere Loss}
In this section, we introduce the proposed loss function. First, we give some denotations and descriptions. The deep face recognition models embeds an image into a $d$-dimensional Euclidean space. $\bm{{x_i}} \in {\mathbb{R}^d}$ denotes the embedding feature of the $i$-th training image, and $y_i$ is the label of $\bm{{x_i}}$. ${\bm{W}}=\left\{ {\bm{{W_1}},\bm{{W_2}}, \ldots ,\bm{{W_C}}} \right\} \in {\mathbb{R}^{d \times C}}$ denotes the weight of the last fully connected layer, where $C$ denotes the number of identities in the training database. ${{\bm{W}_{y_i}}} \in {\mathbb{R}^{d}}$ is seen as the target center feature of identity $y_i$. 

Recent works~\cite{wang2017normface,Liu2017SphereFace,Wang2018CosFace,deng2019arcface} have empirically demonstrated the superiority of constraining deep face features to be discriminative on a hypersphere manifold, where gradients are restricted in the tangent of the hypersphere. We also map deep face features to the hypersphere manifold and optimize cosine similarity to restrict directions of gradients. To help understanding, we illustrate it in Figure~\ref{fig:intro_illu}. With the restricted directions of gradients, the moving directions of samples and target centers are always along the tangent of the hypersphere.

The aim is to decrease the intra-class distance and increase the inter-class distance in a moderate way. Therefore, the sigmoid-constrained hypersphere loss (SFace) of $\bm{{x_i}}$ can be formulated as ${L_{SFace}} = {L_{intra}}\left( {{\theta _{{y_i}}}} \right) + {L_{inter}}\left( {{\theta _j}} \right)$, where $\theta _{{y_i}}$ is the angular distance between ${\bm{{{x_i}}} \mathord{\left/
		{\vphantom {{\bm{x_i}} {\left\| {\bm{x_i}} \right\|}}} \right.
		\kern-\nulldelimiterspace} {\left\| {\bm{x_i}} \right\|}}$ and ${{\bm{{W_{{y_i}}}}} \mathord{\left/
		{\vphantom {{\bm{W_{{y_i}}}} {\left\| {\bm{W_{{y_i}}}} \right\|}}} \right.
		\kern-\nulldelimiterspace} {\left\| {\bm{W_{{y_i}}}} \right\|}}$, and $\theta _{{j}}\,(j \ne {y_i})$ is the angular distance  between ${\bm{{{x_i}}} \mathord{\left/
		{\vphantom {{\bm{x_i}} {\left\| {\bm{x_i}} \right\|}}} \right.
		\kern-\nulldelimiterspace} {\left\| {\bm{x_i}} \right\|}}$ and ${\bm{{{W_{{j}}}}} \mathord{\left/
		{\vphantom {{\bm{W_{{j}}}} {\left\| {\bm{W_{{j}}}} \right\|}}} \right.
		\kern-\nulldelimiterspace} {\left\| {\bm{W_{{j}}}} \right\|}}$. Specifically,  ${L_{intra}}\left( {{\theta _{{y_i}}}} \right)$ and ${L_{inter}}\left( {{\theta _j}} \right)$ are formulated as follows:
\begin{equation}
\label{equ:sface}
\begin{gathered}
L_{intra}\left( {{\theta _{{y_i}}}} \right)=- {[{r_{intra}}\left( {{\theta _{{y_i}}}} \right)]_{b}}\cos \left( {{\theta _{{y_i}}}} \right),\hfill \\
L_{inter}\left( {{\theta _j}} \right)=\sum\nolimits_{j = 1,\;j \ne {y_i}}^C {{[{r_{inter}}\left( {{\theta _j}} \right)]_b}\cos \left( {{\theta _j}} \right)}.
\end{gathered}
\end{equation}In the above equations, $\cos \left( {{\theta _{{y_i}}}} \right) = {{\bm{W_{{y_i}}}^T\bm{x_i}} \mathord{\left/
		{\vphantom {{\bm{W_{{y_i}}}^T\bm{x_i}} {\left\| {\bm{W_{{y_i}}}} \right\|\left\| {\bm{x_i}} \right\|}}} \right.
		\kern-\nulldelimiterspace} {\left\| {\bm{W_{{y_i}}}} \right\|\!\left\| {\bm{x_i}} \right\|}}$, and ${{\cos \left( {{\theta _j}} \right) = \bm{{W_j}}^T\bm{{x_i}}} \mathord{\left/
		{\vphantom {{\cos \left( {{\theta _j}} \right) = {W_j}^T{x_i}} {\left\| {{W_j}} \right\|}}} \right.
		\kern-\nulldelimiterspace} {\left\| {{\bm{W_j}}} \right\|}}\!\left\| {{\bm{x_i}}} \right\|, {\;}j \ne {y_i}$. Since the goal is to obtain precisely control of the optimization degree, we design functions ${r_{intra}}\left( {{\theta _{{y_i}}}} \right)$ and ${r_{inter}}\left( {{\theta _j}} \right)$ to re-scale intra-class and inter-class objectives respectively to further restrict the optimizing speed. ${[\cdot]_b}$ is the block gradient operator, which prevents the contribution of its inputs to be taken into account for computing gradients. In the forward propagation process of SFace, 
\begin{equation}\label{equ:sface_forward}\begin{gathered}L_{SFace}=\hfill \\
-[{{r_{intra}}\left( {{\theta _{{y_i}}}} \right)]_b}\cos \left( {{\theta _{{y_i}}}}\right)\!+\!\sum\nolimits_{j = 1,\!\;j \ne {y_i}}^C \!{{[{r_{inter}}\left( {{\theta _j}} \right)]_b}\cos \left( {{\theta _j}}\right)}.\end{gathered}\end{equation}While in the backward propagation process,
\begin{equation}
\label{equ:sface_back}
\begin{gathered}
\frac{{\partial {L_{SFace}}}}{{\partial\bm{{x_i}}}} =\hfill \\  
-\![{r_{intra}}\!\left( {{{{\theta }}_{{{{y}}_{{i}}}}}} \right)]_b\!\frac{{\partial \cos \left( {{{{\theta }}_{{{{y}}_{{i}}}}}} \right)}}{{\partial {\bm{x_{{i}}}}}} \! + \!\!\mathop \sum \nolimits_{j = 1,\!\;j \ne {y_i}}^C[{r_{inter}}\left( {{\theta _j}} \right)]_b\!\frac{{\partial \cos \left( {{\theta _j}} \right)}}{{\partial \bm{{x_i}}}},\hfill \\
\frac{{\partial {L_{SFace}}}}{{\partial \bm{{W_{{y_i}}}}}}=-[{r_{intra}}\left( {{{{\theta }}_{{{{y}}_{{i}}}}}} \right)]_b\frac{{\partial \cos \left( {{\theta _{{y_i}}}} \right)}}{\partial{\bm{{W_{{y_i}}}}}},\hfill \\
\frac{{\partial {L_{SFace}}}}{{\partial\bm{{W_{{j}}}}}} = [{r_{inter}}\left( {{\theta _j}} \right)]_b\frac{{\partial \cos \left( {{\theta _j}} \right)}}{\partial\bm{{{W_{{j}}}}}},\qquad\qquad\qquad\qquad\qquad
\end{gathered}
\end{equation}where
\begin{equation}
\label{equation:cos}
\begin{gathered}
\frac{{\partial \cos \left( {{\theta _{{y_i}}}} \right)}}{{\partial \bm{x_i}}} = \frac{1}{{\left\| {\bm{x_i}} \right\|}}(\frac{{\bm{W_{{y_i}}}}}{{\left\| {\bm{W_{{y_i}}}} \right\|}} - \cos \left( {{\theta _{{y_i}}}} \right)\frac{{\bm{x_i}}}{{\left\| {\bm{x_i}} \right\|}}),\hfill \\ 
\frac{{\partial \cos \left( {{\theta _j}} \right)}}{{\partial \bm{x_i}}} = \frac{1}{{\left\| {\bm{x_i}} \right\|}}(\frac{{\bm{W_j}}}{{\left\| {\bm{W_j}} \right\|}} - \cos \left( {{\theta _j}} \right)\frac{{\bm{x_i}}}{{\left\| {\bm{x_i}} \right\|}}),\hfill \\ 
\frac{{\partial \cos \left( {{\theta _{{y_i}}}} \right)}}{{\partial \bm{W_{{y_i}}}}} = \frac{1}{{\left\| {\bm{W_{{y_i}}}} \right\|}}(\frac{{\bm{x_i}}}{{\left\| {\bm{x_i}} \right\|}} - \cos \left( {{\theta _{{y_i}}}} \right)\frac{{\bm{W_{{y_i}}}}}{{\left\| {\bm{W_{{y_i}}}} \right\|}}),\hfill \\ 
\frac{{\partial \cos \left( {{\theta _j}} \right)}}{{\partial \bm{W_j}}} = \frac{1}{{\left\| {\bm{W_j}} \right\|}}(\frac{{\bm{x_i}}}{{\left\| {\bm{x_i}} \right\|}} - \cos \left( {{\theta _j}} \right)\frac{{\bm{W_j}}}{{\left\| {\bm{W_j}} \right\|}}).\qquad
\end{gathered}
\end{equation}

\subsection{Gradient Re-scale Function}
\begin{figure}[htbp]
	\centering
	\subfigure[]{
		\begin{minipage}[b]{0.18\textwidth}
			\includegraphics[width=1\textwidth]{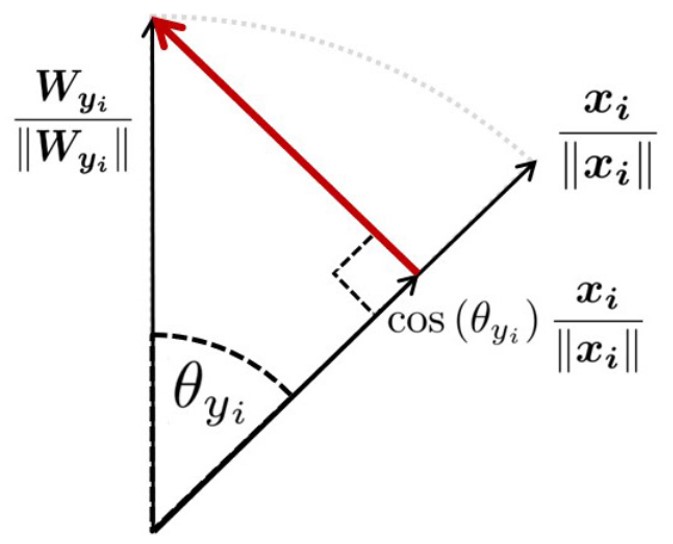}\\
	\end{minipage}}\subfigure[]{
		\begin{minipage}[b]{0.18\textwidth}
			\includegraphics[width=1\textwidth]{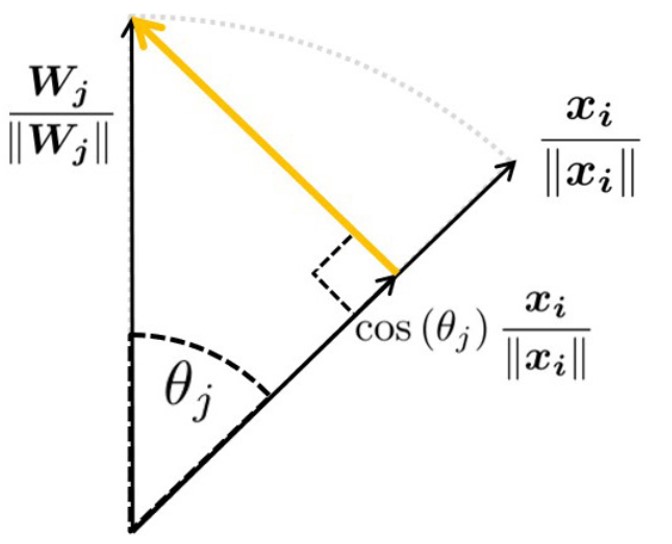} \\
		\end{minipage}
	}
	
	\subfigure[]{
		\begin{minipage}[b]{0.18\textwidth}
			\includegraphics[width=1\textwidth]{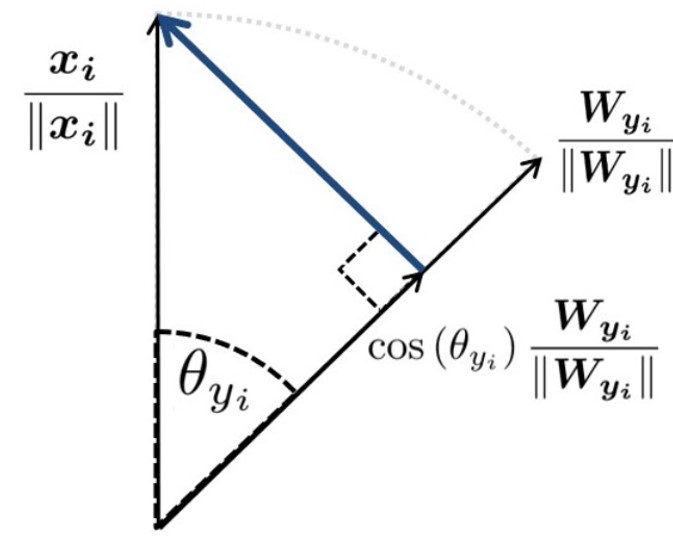}\\
			
		\end{minipage}
	}\subfigure[]{
		\begin{minipage}[b]{0.18\textwidth}
			\includegraphics[width=1\textwidth]{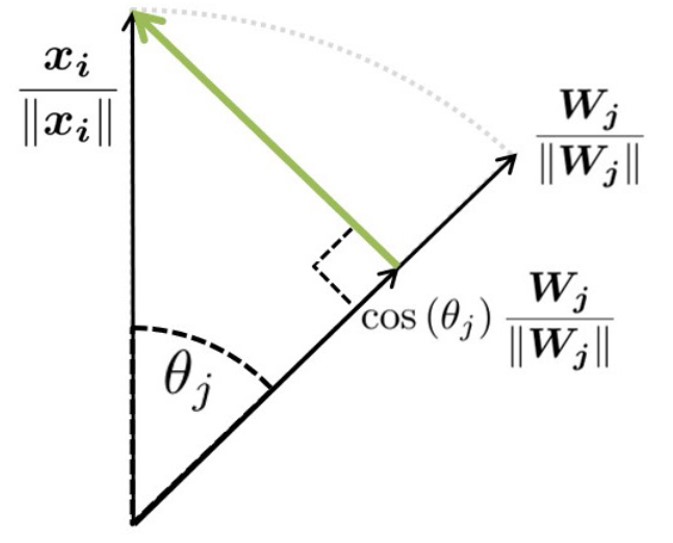} \\
			
		\end{minipage}
	}
	\caption{Illustration of Function~(\ref{equation:cos}), which means that optimization gradients are along the tangent direction. (a)(b)(c)(d) interprets the four orthogonality relationships of Function~(\ref{equation:cos}) respectively.}
	\label{fig:cos}
\end{figure}

The optimization gradients are always along the tangent direction, because $\!\langle\frac{{\partial \cos \left( {{\theta _{{y_i}}}} \right)}}{{\partial \bm{x_i}}},\bm{x_i}\rangle\!=\!0$, $ \langle\frac{{\partial \cos \left( {{\theta _{{j}}}} \right)}}{{\partial \bm{x_i}}},\bm{x_i}\rangle\!=\!0$, $\langle\frac{{\partial \cos \left( {{\theta _{{y_i}}}} \right)}}{{\partial \bm{W_{{y_i}}}}},\bm{W_{{y_i}}}\rangle\!=\!0$, and $\langle{\frac{{\partial \cos \left( {{\theta _j}} \right)}}{{\partial \bm{W_j}}},\bm{W_j}}\rangle\!=\! 0$ (refer to the illustration in Figure~\ref{fig:cos}). In addition, ${\left\| {\bm{x_i}} \right\|}$,${\left\| {\bm{W_{{y_i}}}} \right\|}$ and ${\left\| {\bm{W_{{j}}}} \right\|}$ almost remain unchanged in the training process, for the reason that there are no components of gradients in the radial direction. Function ${r_{intra}}\left( {{\theta _{{y_i}}}}\right)$ and ${r_{inter}}\left( {{\theta _j}} \right)$ are designed to re-scale intra-class and inter-class objectives respectively. These two terms actually re-scale the gradient, \emph{i.e}. control the moving speed of samples and target centers in Figure~\ref{fig:intro_illu}. Therefore we name ${r_{intra}}\left( {{\theta _{{y_i}}}}\right)$ and ${r_{inter}}\left( {{\theta _j}} \right)$ as the gradient re-scale functions. Since the original gradient scales of intra-class and inter-class objectives are proportional to $sin\theta_{y_i}$ and $sin\theta_j$ (refer to Function~(\ref{equation:cos}) and Figure~\ref{fig:cos}), the final gradient scales are proportional to ${v_{intra}}\left( {{\theta _{{y_i}}}} \right)={r_{intra}}\left( {{\theta _{{y_i}}}} \right)sin\theta_{y_i}$ and ${v_{inter}}\left( {{\theta _j}} \right)={r_{inter}}\left( {{\theta _j}} \right)sin\theta_j$.

At the beginning of training, the initial angular distances ${\theta _{{y_i}}}$ and ${\theta _j}$ are all about $\frac{\pi }{2}$. The intra-class loss function decreases ${\theta _{{y_i}}}$ gradually while the inter-class loss function prevents ${\theta _j}$ from being decreased. Therefore, the ideal functions of ${v_{intra}}\left( {{\theta _{{y_i}}}} \right)$ and ${v_{inter}}\left( {{\theta _j}} \right)$ should satisfy at least three properties as follows: (1) The function ${v_{intra}}\left( {{\theta _{{y_i}}}} \right)$ should be non-negative and monotonically increasing on the interval $[0,\frac{\pi }{2}]$, so that the moving speed of $\bm{{x_i}}$ and ${\bm{W_{{y_i}}}}$ decreases gradually as they approaching to each other. (2) The function ${v_{inter}}\left( {{\theta _{j}}} \right)$ should be non-negative on the interval $[0,\frac{\pi }{2}]$, so that the moving speed of $\bm{{x_i}}$ and ${\bm{W_{{j}}}}$ increases rapidly as they start approaching to each other. (3) Considering the imperfection of training databases, there should be two flexible intervals to suppress the moving speed, one is around ${\theta _{{y_i}}}\approx0$ of ${v_{intra}}\left( {{\theta _{{y_i}}}} \right)$ and the other is around ${\theta_{j}}\approx\frac{\pi }{2}$ of ${v_{inter}}\left( {{\theta _{j}}} \right)$, so that both intra-class and inter-class objectives can be optimized with a moderate target rather than be minimized or maximized strictly.

\begin{figure}[htbp]
	\centering
	\subfigure[]{
		\begin{minipage}[b]{0.475\textwidth}
			\includegraphics[width=1\textwidth]{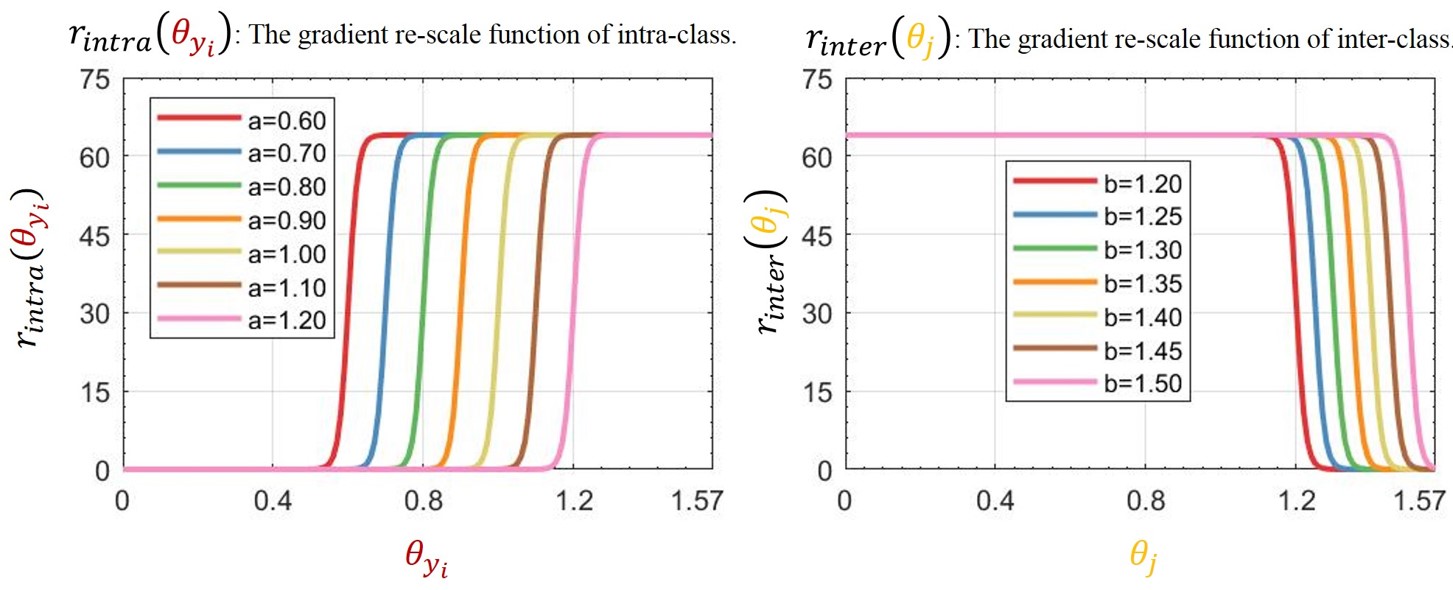} \\
		\end{minipage}
	}

	\subfigure[]{
		\begin{minipage}[b]{0.475\textwidth}
			\includegraphics[width=1\textwidth]{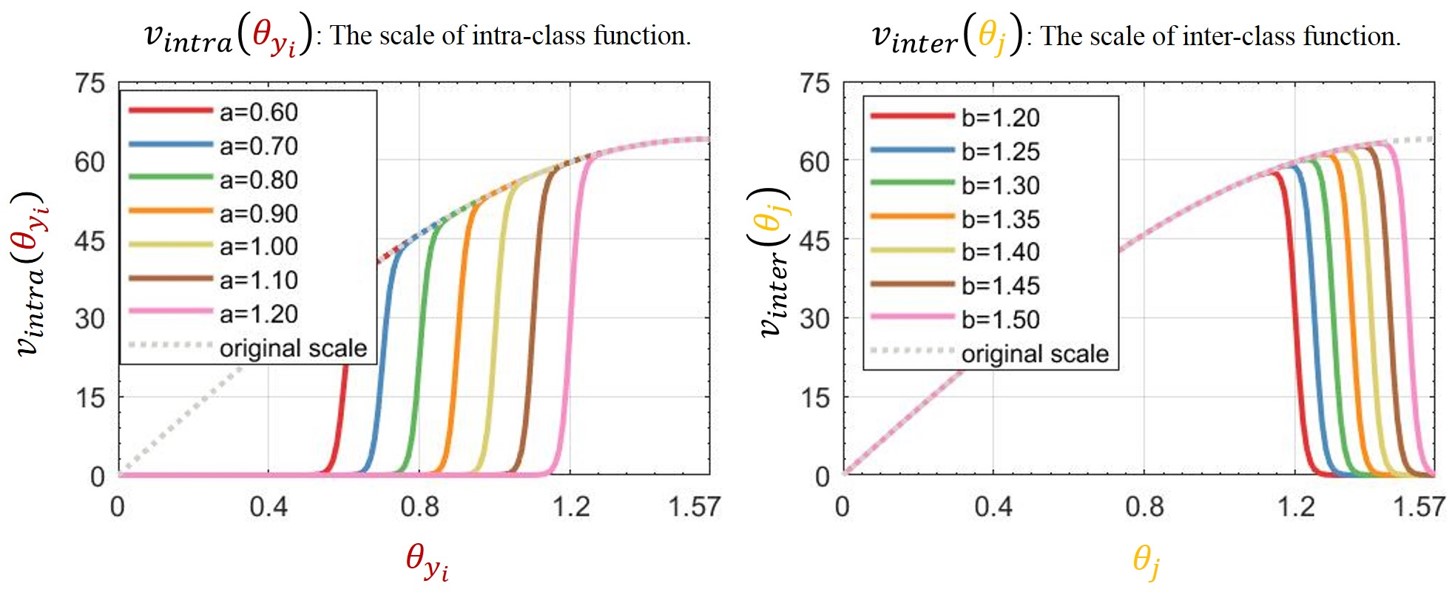} \\

		\end{minipage}
	}
\caption{(a) The sigmoid curves of intra-class gradient re-scale function ${r_{intra}}\left( {{\theta _{{y_i}}}} \right)$ and inter-class gradient re-scale function ${r_{inter}}\left( {{\theta _j}} \right)$ of SFace. (b) The final scale curves of intra-class gradient ${v_{intra}}\left( {{\theta _{{y_i}}}} \right)$ and inter-class gradient ${v_{inter}}\left( {{\theta _j}} \right)$ of SFace.}
\label{fig:method_curve}
\end{figure}

Eventually, we choose sigmoid functions as the gradient re-scale functions. The specific forms are,
\begin{equation}
\label{equa:rescale}
\begin{gathered}
\quad{r_{intra}}\left( {{\theta _{{y_i}}}} \right) = \frac{s}{{1 + {e^{ - k*\left( {{\theta _{{y_i}}} - a} \right)}}}}, \hfill \\
{r_{inter}}\left( {{\theta _j}} \right) = \frac{s}{{1 + {e^{k*\left( {{\theta _{j}} - b} \right)}}}}.
\end{gathered}
\end{equation}$s$ is the upper asymptote of two sigmoid curves as the initial scale of gradient, and $k$ is the control the slope of sigmoid curves. Hyperparameters $a$ and $b$ decide the horizontal intercept of two sigmoid curves and actually control the flexible interval to suppress the moving speed. Therefore $a$ and $b$ are vital parameters should be selected according to characteristics of a specific training database, which we will discuss later. The sigmoid curve functions of ${r_{intra}}\left( {{\theta _{{y_i}}}} \right)$ and ${r_{inter}}\left( {{\theta _j}} \right)$ are illustrated in (a) of Figure~\ref{fig:method_curve}. With the gradient re-scale functions, scales of intra-class gradient and inter-class gradient in theory are proportional to ${v_{intra}}\left( {{\theta _{{y_i}}}} \right)={r_{intra}}\left( {{\theta _{{y_i}}}} \right)sin\theta_{y_i}$ and ${v_{inter}}\left( {{\theta _j}} \right)={r_{inter}}\left( {{\theta _j}} \right)sin\theta_j$, shown in (b) of Figure~\ref{fig:method_curve}. The entire training process of SFace is summarized in Algorithm~\ref{alg:1}, which is easy for implementation.

\begin{algorithm}[htbp]
	\caption{SFace}
	\label{alg:1}
	\LinesNumbered
	\KwIn{Embedding feature $\bm{x_i}$ with label $y_i$, parameters of the embedding network $\varTheta$, parameters of the last fully-connected layer ${\bm{W}}$ (composed of $\bm{{W_{y_i}}}$ and $\bm{{W_{j}}}$ ($j\ne{y_i}$)), SFace parameters $s$ and $k$, $a$ and $b$, the number of iteration $i = 0$, learning rate $\lambda^{(i)}$, }	
	\While{not converged}{
		$i = i+1$\;
		Compute the intra-distance by $\theta_{y_i} = \arccos \left( {{\bm{W_{{y_i}}}^T\bm{x_i}} \mathord{\left/
				{\vphantom {{\bm{W_{{y_i}}}^T\bm{x_i}} {\left\| {\bm{W_{{y_i}}}} \right\|\left\| {\bm{x_i}} \right\|}}} \right.
				\kern-\nulldelimiterspace} {\left\| {\bm{W_{{y_i}}}} \right\|\!\left\| {\bm{x_i}} \right\|}} \right)$\; 
		Compute the inter-distance by $\theta_{j} =\arccos \left( {\bm{{W_j}}^T\bm{{x_i}}} \mathord{\left/
			{\vphantom {{\cos \left( {{\theta _j}} \right) = {W_j}^T{x_i}} {\left\| {{W_j}} \right\|}}} \right.
			\kern-\nulldelimiterspace} {\left\| {{\bm{W_j}}} \right\|}\!\left\| {{\bm{x_i}}} \right\|
		\right), {\;}j \ne {y_i}$\;
		Compute gradient re-scale functions by Equation~\ref{equa:rescale}\;
		Compute the loss by Equation~\ref{equ:sface_forward}\;
		Compute the gradients of $\bm{x_i}$ and ${\bm{W}}$ by Equation~\ref{equ:sface_back}\;
		Update parameters ${\bm{W}}$ and $\varTheta$ by 
		${\bm{W}}={\bm{W}}-\lambda ^{\left( i \right)}\frac{\partial L_{SFace}}{\partial {\bm{W}}},$
		$\varTheta =\varTheta -\lambda ^{\left( i \right)}\frac{\partial L_{SFace}}{\partial \bm{x_i}}\frac{\partial \bm{x_i}}{\partial \varTheta}$\;
		
	}
	\KwOut{${\bm{W}}$, $\varTheta$.}	
\end{algorithm}

\subsection{Relation to Softmax Based Loss}
\label{subsection:relation}
We have mentioned in Section~\ref{sec:related} that SFace in form can be categorized as the metric learning method, but it has a close connection to the softmax based loss functions. In this section, we discuss this relation in details. 

We start from the original softmax loss function. For each embedding feature $\bm{{x_i}}$, the softmax loss can be formulated as:
\begin{equation}
L =  - \log {P_{y_i}} = -  \log  \frac{{{e^{\bm{{W_{{y_i}}}}^T\bm{{x_i}} + \bm{{b_{{y_i}}}}}}}}{{\sum\nolimits_{j = 1}^C {{e^{\bm{{W_j}}^T\bm{{x_i}} + \bm{{b_j}}}}} }}.
\end{equation}
$\bm{{x_i}} \in {\mathbb{R}^d}$ denotes the embedding feature of the $i$-th training image, and $y_i$ is the label of $\bm{{x_i}}$. ${P_{y_i}}$ is the predicted probability of assigning $\bm{{x_i}}$ to class $y_i$. $C$ is the number of identities, ${{\bm{W}_{j}}} \in {\mathbb{R}^{d}}$ is the $j$-th column of the weight of the last fully connected layer, $\bm{b_j}\in {\mathbb{R}^{C}}$ is the bias. Softmax based loss functions~\cite{wang2017normface,Liu2017SphereFace,Wang2018CosFace,deng2019arcface} remove the bias term and transform $\bm{{W_j}}^T\bm{{x_i}} = s\cos {\theta_j}$. To further improve the performance, large margin is adopted in the $\cos{\theta_{y_i}}$ term~\cite{Liu2017SphereFace,Wang2018CosFace,deng2019arcface}. Therefore, softmax based loss functions can be formulated as:
\begin{equation}
\label{equation:osoft}
\begin{gathered}
L =  - \log {P_{y_i}} = - \log \frac{{{e^{sf(\theta _{{y_i}})}}}}{{{e^{sf(\theta _{{y_i}})}} + \sum\nolimits_{j = 1,j \ne {y_i}}^C {{e^{s\cos {\theta _j}}}} }}, 
\end{gathered}\end{equation}where $f(\theta _{{y_i}}) = \cos {\theta _{{y_i}}}$ in NSoftmax~\cite{wang2017normface}, $f(\theta _{{y_i}}) = \cos {\theta _{{y_i}}} - m$ in CosFace~\cite{Wang2018CosFace}, and $f(\theta _{{y_i}}) = \cos ({\theta _{{y_i}}} + m)$ in ArcFace~\cite{deng2019arcface}. With the influence of the loss function, $\theta _{{y_i}}$ is decreased and $\theta_j$ is increased in theory. In the backward propagation process, 
\begin{equation}
\label{equation:soft}
\begin{gathered}
\frac{{\partial L}}{{\partial \cos {\theta _{{y_i}}}}} = s({P_{{y_i}}} - 1)\frac{{\partial f({\theta _{{y_i}}})}}{{\partial \cos {\theta _{{y_i}}}}} \hfill \\
\qquad\quad\quad=-\frac{s{\sum\nolimits_{j = 1,j \ne {y_i}}^C {{e^{s\cos {\theta _j}}}} }}{{{e^{sf({\theta _{{y_i}}})}} + \sum\nolimits_{j = 1,j \ne {y_i}}^C {{e^{s\cos {\theta _j}}}} }}\frac{{\partial f({\theta _{{y_i}}})}}{{\partial \cos {\theta _{{y_i}}}}},\\
\frac{{\partial L}}{{\partial \cos {\theta _j}}} = s{P_j}
= \frac{{s{e^{s\cos {\theta _j}}}}}{{{e^{sf( {\theta _{{y_i}}})}} + \sum\nolimits_{k = 1,k \ne {y_i}}^C {{e^{s\cos {\theta _k}}}} }},\hfill\\ 
\end{gathered}
\end{equation}
where $\frac{{\partial f({\theta _{{y_i}}})}}{{\partial \cos {\theta _{{y_i}}}}}=1$ in NSoftmax~\cite{wang2017normface} and CosFace~\cite{Wang2018CosFace}, and $\frac{{\partial f({\theta _{{y_i}}})}}{{\partial \cos {\theta _{{y_i}}}}} = \frac{{\sin ({\theta _{{y_i}}} + m)}}{{\sin {\theta _{{y_i}}}}}$ in ArcFace~\cite{deng2019arcface}.

\begin{figure}[htbp]
	\centering
	\subfigure[NSoftmax~\cite{wang2017normface}]{
		\begin{minipage}[b]{0.475\textwidth}
			\includegraphics[width=1\textwidth]{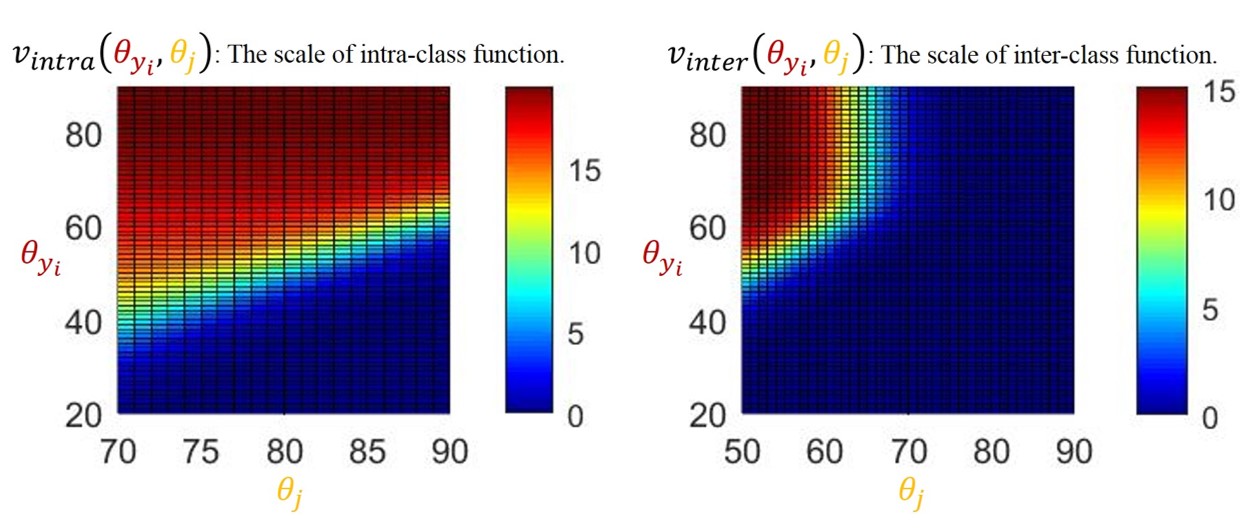} \\
		\end{minipage}
	}
	
	\subfigure[CosFace~\cite{Wang2018CosFace}]{
		\begin{minipage}[b]{0.475\textwidth}
			\includegraphics[width=1\textwidth]{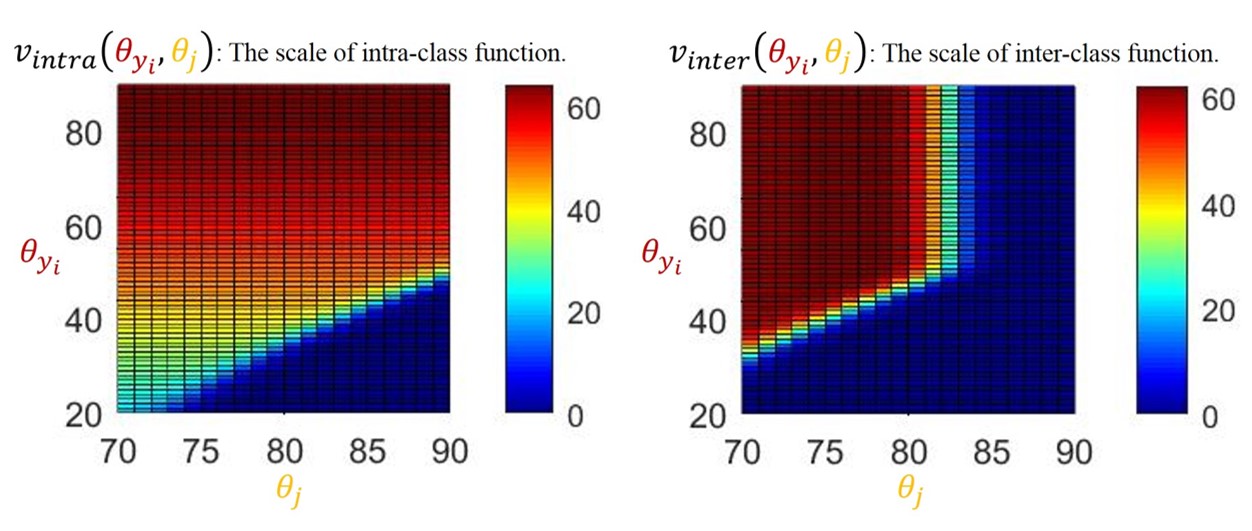} \\
		\end{minipage}
	}
	
	\subfigure[ArcFace~\cite{deng2019arcface}]{
		\begin{minipage}[b]{0.475\textwidth}
			\includegraphics[width=1\textwidth]{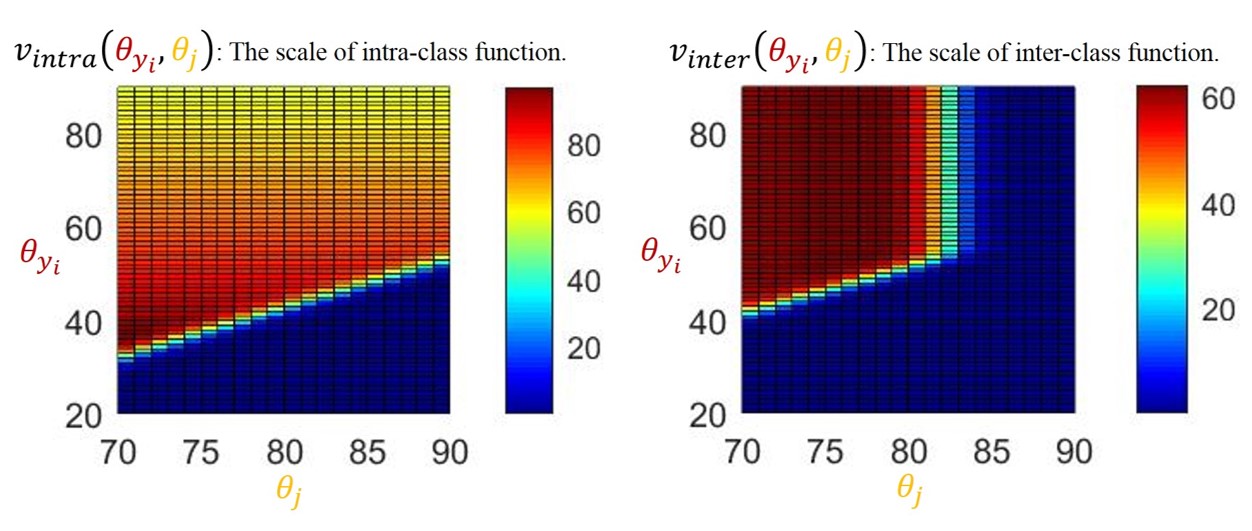} \\
			
		\end{minipage}
	}
	\caption{Under some ideal assumptions, the scale of intra-class gradient ${v_{intra}}\left( {{\theta _{{y_i}}}},{{\theta _{{j}}}}\right)$ and inter-class gradient ${v_{inter}}\left({{\theta _{{y_i}}}}, {{\theta _j}} \right)$ of (a) NSoftmax~\cite{wang2017normface}, (b) CosFace~\cite{Wang2018CosFace}, and (c) ArcFace~\cite{deng2019arcface}.  Softmax based loss functions~\cite{wang2017normface,Wang2018CosFace,deng2019arcface} can be understand as a kind of special metric learning method with specific speed constraints decided by the intra-class distance ${\theta_{y_i}}$ and the inter-class distances $\theta _{{j}}\,(j \ne {y_i})$.}
	\label{fig:soft_method_curves}
\end{figure}

Further, the softmax based functions are equivalent to the following loss functions for training face models
	\begin{equation}
	\begin{gathered}
	\label{equation:equi}
	L = 
	-[{{r_{intra}}\left( {{\theta _{{y_i}}}},{{\theta _{j}}} \right)]_b}\cos \left( {{\theta _{{y_i}}}} \right)+ \hfill \\
\quad\qquad\qquad\qquad \sum\nolimits_{j = 1,\;j \ne {y_i}}^C {{[{r_{inter}}\left({{\theta _{{y_i}}}},{{\theta _j}}\right)]_b}\cos\left( {{\theta _j}}\right)},
\end{gathered}\end{equation}where gradient re-scale functions are, \begin{equation}\label{eq:soft_intra_rescale}
{r_{intra}}\left( {{\theta _{{y_i}}}},{{\theta _{j}}} \right)=\frac{{s\sum\nolimits_{j = 1,j \ne {y_i}}^C {{e^{s\cos {\theta _j}}}} }}{{{e^{sf({\theta _{{y_i}}})}} + \sum\nolimits_{j = 1,j \ne {y_i}}^C {{e^{s\cos {\theta _j}}}} }}\frac{{\partial f({\theta _{{y_i}}})}}{{\partial \cos {\theta _{{y_i}}}}},\end{equation} and \begin{equation}\label{eq:soft_inter_rescale}
{r_{inter}}\left( {{\theta _{{y_i}}}}, {{\theta _j}} \right)=\frac{{s{e^{s\cos {\theta _j}}}}}{{{e^{sf({\theta _{{y_i}}})}} + \sum\nolimits_{k = 1,k \ne {y_i}}^C {{e^{s\cos {\theta _k}}}} }}.\end{equation} Since only backward propagation have influence on the network parameters of deep face recognition models, and the backward propagation function (\ref{equation:soft}) of softmax based loss functions and function (\ref{equation:equi}) are the same. Therefore loss function (\ref{equation:osoft}) are equivalent to loss function (\ref{equation:equi}) in the training process. 

Now from equations (\ref{equation:soft})(\ref{equation:equi})(\ref{eq:soft_intra_rescale})(\ref{eq:soft_inter_rescale}), we can see that softmax based loss functions can be understood as a kind of special metric learning method with the speed constraints on a hypersphere. However, both the gradient re-scale functions (speed constraints) of intra-class and inter-class are decided by the intra-class distance ${\theta_{y_i}}$ and the inter-class distances $\theta _{{j}}\,(j \ne {y_i})$. To better understanding of the optimization of softmax based loss functions, we hypothesize that all the inter-class distances $\theta _{{j}}\,(j \ne {y_i})$ are the same ideally, and plot scale curves of the intra-class gradient ${v_{intra}}\left( {{\theta _{{y_i}}}},{{\theta _j}} \right)={r_{intra}}\left( {{\theta _{{y_i}}}},{{\theta _j}} \right)sin\theta_{y_i}$ and the inter-class gradient ${v_{inter}}\left( {{\theta _{{y_i}}}}, {{\theta _j}} \right)={r_{inter}}\left( {{\theta _{{y_i}}}}, {{\theta _j}} \right)sin\theta_j$ of (a) NSoftmax~\cite{wang2017normface}, (b) CosFace~\cite{Wang2018CosFace}, and (c) ArcFace~\cite{deng2019arcface} in Figure~\ref{fig:soft_method_curves}. At the beginning of training, the intra-class distance $\theta_{y_i}$ and inter-class distances $\theta_j$ is about 90 degrees ($\frac{\pi}{2}$). We can see that, from the intra-class sub-figure (left) of Figure~\ref{fig:soft_method_curves}, with the high intra-class gradient ${v_{intra}}$, the intra-class distance $\theta_{y_i}$ will decrease gradually. While at the same time, as the intra-class distance $\theta_{y_i}$ decreases, from the inter-class sub-figure (right) of Figure~\ref{fig:soft_method_curves}, the inter-class gradient ${v_{inter}}$ will decrease, which will relax the inter-class constraints and decrease the inter-class distance $\theta_j$. Then, we come back to the intra-class sub-figure (left) of Figure~\ref{fig:soft_method_curves}, as the inter-class distances $\theta_j$ decrease, the change curve of intra-class gradient ${v_{intra}}$ $vs$ $\theta_{y_i}$ will also changed. 

In the optimization of softmax based loss, the intra-class and inter-class distance will always have influence on each other. Therefore, in conclusion, softmax based loss functions actually lack the ability to control intra-class and inter-class optimizations precisely. However, compared with softmax based loss functions, both intra-class and inter-class distance of SFace (Figure~\ref{fig:method_curve}) can be constrained to a designed degree therefore can be optimized in a moderate way, which is exactly the advantage of SFace.   

\section{Experiments}
\label{sec:exp}
\subsection{Experimental settings}
We separately train models on training databases including CASIA-WebFace~\cite{Yi2014CASIA}, VGGFace2~\cite{Cao18}, MS1MV2~\cite{guo2016msceleb} databases, which have been elaborated by semi-automatic data cleaning algorithms, to evaluate our methods and conduct fair comparison with state-of-the-art loss functions. The compared loss functions include softmax, NSoftmax~\cite{wang2017normface}, SphereFace~\cite{Liu2017SphereFace}, CosFace~\cite{Wang2018CosFace}, ArcFace~\cite{deng2019arcface}, Combined loss~\cite{deng2019arcface}, D-softmax~\cite{he2019softmax} and so on.

\textbf{Evaluation Databases.} We evaluate on LFW~\cite{LFWTech}, YTF~\cite{Wolf2011Face}, CFP-FP~\cite{sengupta2016frontal}, AgeDB-30~\cite{moschoglou2017agedb}, CALFW~\cite{zheng2017CALFW}, CPLFW~\cite{CPLFWTech}, MegaFace~\cite{kemelmacher2016megaface}, IJB-A~\cite{klare2015pushing} and IJB-C~\cite{maze2018iarpa} databases. 

LFW~\cite{LFWTech} database contains 13,233 face images from 5,749 different identities. YTF~\cite{Wolf2011Face} is a database of face video collected from YouTube, which consists of 3,425 videos of 1,595 different people. CFP-FP database~\cite{sengupta2016frontal} is built for facilitating large pose variation in unconstrained settings. AgeDB-30 database~\cite{moschoglou2017agedb} is a manually collected cross-age database in unconstrained settings. Cross-Age LFW (CALFW)~\cite{zheng2017CALFW} and Cross-Pose LFW (CPLFW)~\cite{CPLFWTech} databases are constructed based on LFW database, to emphasize cross-age challenge and cross-pose challenge in face recognition. 

MegaFace~\cite{kemelmacher2016megaface} is a large public available testing benchmark, which evaluates the performance of face models at the million scale distractors. We use FaceScrub database~\cite{Ng2015A} as the probe set, which contains 106,863 images from 530 celebrities. The gallery set is a subset of Flickr photos and it consists of more than one million images. Recently, research~\cite{deng2019arcface} points out that there are many wrong labels in the MegaFace database and the noise significantly affects the performance. Therefore, for comparison, in this paper we report experimental results on both the original MegaFace database and the refined version~\cite{deng2019arcface}.

IJB-A~\cite{klare2015pushing} and IJB-C~\cite{maze2018iarpa} databases address the unconstrained face recognition, which contain both still images and video frames. IJB-A database contains 500 subjects with 5,396 still images and 20,395 video frames. IJB-C database further increases emphasis on occlusion and diversity of subject occupation and geographic origin population, containing 3,531 subjects with 31.3K still images and 117.5K frames from 11,779 videos. We evaluate the models on the standard verification setting (matching between the Mixed Media probes and two galleries) and identification protocol (1:N Mixed Media probes across two galleries).

\textbf{Training and Testing.} We use MxNet~\cite{Chen2015MXNet} to implement all the experiments. For the fair comparison, the CNN architecture used in our work is the same ResNet~\cite{he2016deep} networks as~\cite{deng2019arcface}, which applies the ``BN~\cite{ioffe2015batch}-Dropout~\cite{Srivastava2014Dropout}-FC-BN'' structure to get the final 512-$D$ embedding feature. The data preprocessing follows settings of insightface~\cite{deng2019arcface}. That is, horizontally flip with a probability of 50\% is used for training data augmentation. In addition, all the images are normalized by subtracting 127.5 and dividing by 128. All the models are trained with stochastic gradient descent (SGD) algorithm from scratch. Models trained on CASIA-WebFace database are trained on 2 GPUs and the total batch size is 256. The learning rate is started from 0.1 and divided by 10 at the 100k, 140k, 160k iterations. Models trained on MS1MV2 database are trained on 4 GPUs and the total batch size is 512. The learning rate is started from 0.1 and divided by 10 at the 100k, 160k, 220k iterations. Models trained on VGGFace2 database are trained on 4 GPUs and the total batch size is 512. The learning rate is started from 0.1 and divided by 10 at the 80k, 100k, 160k iterations. The parameter $s$ for SFace is set to 64, $k$ is set to 80. The intra-class and inter-class parameters $a$ and $b$ control the optimization and should be decided according to specific training databases, which will be introduced later.

\begin{table*}
	\renewcommand\arraystretch{1.05}
	\begin{center}
		\caption{Comparison of different loss functions with SFace. Models are trained on CASIA-WebFace~\cite{Yi2014CASIA} using ResNet50. The combined loss~\cite{deng2019arcface} adopts the combined margin $\cos \left( {{m_1}\theta  + {m_2}} \right) - {m_3}$. The evaluation benchmark contains IJB-C~\cite{maze2018iarpa} (TAR@FAR=1e-5,1e-4,1e-3), YTF~\cite{Wolf2011Face} (\%) databases, and average performance (\%) on LFW~\cite{LFWTech}, CFP-FP~\cite{sengupta2016frontal}, AgeDB-30~\cite{moschoglou2017agedb}, CALFW~\cite{zheng2017CALFW} and CPLFW~\cite{CPLFWTech} databases.}
		\label{table:table1}
		\scalebox{0.984}{
			\begin{tabular}{|c|c|c|c|c|c|c|c|c|c|c|}
				\hline
				\multirow{2}{*}{Method} & \multicolumn{3}{c|}{IJB-C} & \multicolumn{1}{c|}{\multirow{2}{*}{YTF}} & \multirow{2}{*}{Avg.} & \multicolumn{1}{c|}{\multirow{2}{*}{LFW}} & \multicolumn{1}{c|}{\multirow{2}{*}{CFP-FP}} & \multicolumn{1}{c|}{\multirow{2}{*}{AgeDB-30}} & \multicolumn{1}{c|}{\multirow{2}{*}{CPFLW}} & \multicolumn{1}{c|}{\multirow{2}{*}{CALFW}} \\ \cline{2-4}
				& FAR=1e-5    & FAR=1e-4    & FAR=1e-3   & \multicolumn{1}{c|}{}                     &\multicolumn{1}{c|}{}       & \multicolumn{1}{c|}{}                     & \multicolumn{1}{c|}{}                        & \multicolumn{1}{c|}{}                          & \multicolumn{1}{c|}{}                       & \multicolumn{1}{c|}{}                       \\ \hline\hline
				softmax&64.57&77.57&88.03&95.60
				&93.82&99.25&95.10&93.28&88.97&92.48\\ 
				NSoftmax~\cite{wang2017normface} (s=20.0)&67.82&79.88&89.33&95.54
				&93.72&99.23&95.00&93.17&88.82&92.40\\ 
				SphereFace~\cite{Liu2017SphereFace} (m=1.35)&46.73&61.54&76.10&93.18
				&92.99&99.17&94.76&92.60&86.50&91.93\\ 
				CosFace~\cite{Wang2018CosFace} (m=0.35)&75.58&85.03&92.00&95.76
				&94.91&99.53&95.50&95.23&90.32&93.97\\
			    ArcFace~\cite{deng2019arcface} (m=0.3)& 73.55 & 84.60 & 91.90 & 95.80              & 94.65               & \textbf{99.57}     & 95.26                 & 94.40                   & 90.10                & 93.93       \\ 
				ArcFace~\cite{deng2019arcface} (m=0.4)& 72.49& 83.76& 91.21& 96.06& 94.91& 99.52& 95.76& 95.00&90.43& 93.87\\ 
				ArcFace~\cite{deng2019arcface} (m=0.5)&70.15&81.48&90.26&95.66
				&94.83&99.52&95.60&\bf{95.30}&89.97&93.77\\ 
				Combined~\cite{deng2019arcface} (m = 0.9,0.4,0.15)&73.99&83.91&91.63&95.86
				&94.90&99.48&95.56&94.97&90.68&93.82\\ 
				D-softmax~\cite{he2019softmax} (d=0.9)&71.48&83.56&91.23&95.42
				&94.29&99.50&95.44&93.95&89.60&92.95\\ \hline\hline
				SFace (a=0.87, b=1.20)&77.13&86.38&92.52&95.82
				&\bf{94.93}&99.50&\bf{95.81}&95.10&90.18&\bf{94.07}\\
				SFace (a=0.90, b=1.20)
				&76.77&85.95&92.37&95.86
				&94.88&\bf{99.57}&95.67&95.00&90.22&93.95\\ 
				SFace (a=0.93, b=1.20)
				&\bf{77.77}&86.38&92.52&\bf{96.08}
				&94.88&99.48&95.81&94.87&90.28&93.97\\ 			
				SFace (a=0.90, b=1.30)&76.92&\bf{87.27}&\bf{93.11}&96.00&94.80
				&99.57&95.26&94.82&\bf{90.68}&93.70\\ \hline				
			\end{tabular}
		}
	\end{center}
	
\end{table*}

\subsection{Experiment on the CASIA-WebFace Database}
CASIA-WebFace database~\cite{Yi2014CASIA} contains 0.49M images from 10,575 celebrities, which is the first widely used large training database in deep face recognition. While recently it has been seen as a relatively small-scale database compared with other Million-scale ones~\cite{guo2016msceleb,Cao18}. According to the research~\cite{wang2018devil}, there are 9.3\%-13.0\% label noise in CASIA-WebFace database. That is, the original CASIA-WebFace database is exactly the database using semi-automatic annotation with low level noise. We use the arcface version~\cite{deng2019arcface} with 0.49M images from 10,572 identities. We first implement our method on it and compare with the state-of-art loss functions. Then, we experiment on the noise-controlled WebFace database to further evaluate our method under training databases with different noise levels, and study the choice of hyper-parameters.

\subsubsection{Experiment on the CASIA-WebFace Database}
We train face models on CASIA-WebFace database supervised by softmax, NSoftmax~\cite{wang2017normface}, SphereFace~\cite{Liu2017SphereFace}, CosFace~\cite{Wang2018CosFace}, ArcFace~\cite{deng2019arcface}, Combined loss~\cite{deng2019arcface} with combined margin $\cos \left( {{m_1}\theta  + {m_2}} \right) - {m_3}$, D-softmax~\cite{he2019softmax}, and SFace respectively. The source codes of most compared methods can be downloaded from the github. In addition, we implement D-softmax~\cite{he2019softmax} by ourselves. Since the performance of all the above loss functions is sensitive to the choice of hyper-parameters, we list them in the Table~\ref{table:table1}, which are determined according to the suggestion. All the models are trained on the ResNet50 which we have mentioned above. For SFace, we choose intra-class and inter-class hyper-parameters $a$ and $b$ by taking reference to the experience of the final models of large margin loss functions, and then tuning them. In the experiment, both intra-class and inter-class parameters have crucial influence. Table~\ref{table:table1} lists the experimental results, our method is compared with the recent advanced loss functions. As shown, under the same training and test settings, our method significantly improves the results on several evaluation benchmarks, especially TAR at very low FAR on the well-known challenging IJB-C database, which demonstrates the superiority of our method on a semi-automatic annotated face training database with low level noise. 

From Table~\ref{table:table1}, we select three models trained supervised by SFace and two classic methods, NSoftmax and ArcFace, respectively, and analyze these models. We extract the deep features of images in the training database, and calculate intra-class and inter-class angles (distances) statistics. Specifically, using the manual refined image list~\cite{rwebface_wang} released by~\cite{wang2017normface}, we can split the training database (0.49M images) into clean images (0.45M) and label noises (0.04M). Therefore, the mean angles (distances) between embedding feature $x_i$ and the embedding feature $W_{y_i}$ of clean images and label noise can be calculated respectively. In addition, we calculate the mean angles between different $W_j$. The results are listed in Table~\ref{table:ana}. We can see that, compared with NSoftmax and SFace, ArcFace optimizes training samples in a more strict way. That is, the intra-class angles (distances) of ArcFace are smaller. The decrease of intra-class angles (distances) of clean images is a good trend. However, the intra-class angles (distances) of label noise are also decreased, which is not a good phenomenon. While SFace keep a better balance between decreasing the intra-class angles (distances) and preventing overfitting to label noise. The reason may be that with the precisely control to a cutoff point, the clean training samples are optimized earlier and more easily, while the label noise can be left behind to prevent them close to the wrong labeled targets. At the same time, the inter-class class optimization guarantees that different identities still remain to be orthogonal to each other. 

To evaluate the proposed gradient re-scale function of SFace, we compare face models trained on loss function~(\ref{equ:sface}) with three different gradient re-scale functions: constant value (no gradient re-scale), the piecewise functions, and the sigmoid functions (SFace). Specifically, the piecewise function can be seen as the ``steep version'' of the sigmoid functions, formulated as follows, 
\begin{equation}
\begin{gathered}
{r_{intra}}\left( {{\theta_{{y_i}}}} \right) = s*sign\left( \max \left( \theta _{y_i}-a,0 \right) \right)  , \hfill \\
{r_{inter}}\left( {{\theta _j}} \right) =s*sign\left( \max \left( b-\theta _j,0 \right) \right),
\end{gathered}
\end{equation}where $sign(*)$ is the sign function to extract the sign of a real number. For the piecewise and sigmoid functions, the hyper-parameters $a$, $b$ are set as the same, 0.9 and 1.3. The experimental results are listed in Table~\ref{table:scale_gradient}. We can see that, the proposed sigmoid gradient re-scale function has better performance than the constant value and the piecewise version. 

\begin{table}[htbp]
	\renewcommand\arraystretch{1.02}
	\begin{center}
		\caption{The angles (distances) statistics under different loss functions (NSoftmax~\cite{wang2017normface}, ArcFace~\cite{deng2019arcface} and SFace models trained on WebFace database (0.49M images)). Each column denotes one loss function. ``Clean-Intra" and ``Noise-Intra" refers to calculate the mean angles (distances) between embedding feature $x_i$ and the embedding feature $W_{y_i}$ of clean images and label noise, respectively. We use the manual refined image list released by~\cite{rwebface_wang} to split the training database (0.49M images) into clean images (0.45M) and label noises (0.04M). ``Delta-Intra" is the difference between ``Noise-Intra" and  ``Clean-Intra". ``Inter" refers to the mean angles between different $W_j$. }
		\label{table:ana}
		\scalebox{1.0}{
			\begin{tabular}{|c|c|c|c|}
				\hline
				&NSoftmax~\cite{wang2017normface}&ArcFace~\cite{deng2019arcface}&SFace\\
				\hline\hline
				Clean-Intra&44.42&35.31&39.68\\\hline
				Noise-Intra&50.85&40.09&47.30\\\hline
				Delta-Intra&6.43&4.78&7.62\\\hline
				Inter&89.75$\pm$5.55&89.99$\pm$4.73&89.96$\pm$4.67\\
				\hline
			\end{tabular}
		}
	\end{center}
	
\end{table}

\begin{table}[htbp]
	\renewcommand\arraystretch{1.1}
	\begin{center}
		\caption{Comparison of three different gradient re-scale functions: constant value (no gradient re-scale), the piecewise functions, and the sigmoid functions (SFace). Models are trained on CASIA-WebFace~\cite{Yi2014CASIA} using ResNet50. The average performance (\%) on LFW~\cite{LFWTech}, CFP-FP~\cite{sengupta2016frontal}, AgeDB-30~\cite{moschoglou2017agedb}, CALFW~\cite{zheng2017CALFW} and CPLFW~\cite{CPLFWTech} databases is used for evaluation.}
		\label{table:scale_gradient}
		\scalebox{0.91}{
			\begin{tabular}{|c|c|c|c|c|c|c|}
				\hline
				Method      & Avg.    & LFW     & CFP-FP  & AgeDB-30 & CPLFW   & CALFW   \\ \hline\hline
				Constant & 90.05 & 98.30 & 90.46 & 89.55  & 83.15 & 88.78\\ \hline
				Piecewise & 94.64 & 99.45 & 94.90 & 94.73  & 90.08 & \bf{94.03} \\ \hline
				Sigmoid & \bf{94.80} & \bf{99.57} & \bf{95.26} & \bf{94.82}  & \bf{90.68} & 93.70 \\ \hline
			\end{tabular}
		}
	\end{center}
\end{table}

\begin{figure}[htbp]
	\center
	\includegraphics[width=1\linewidth]{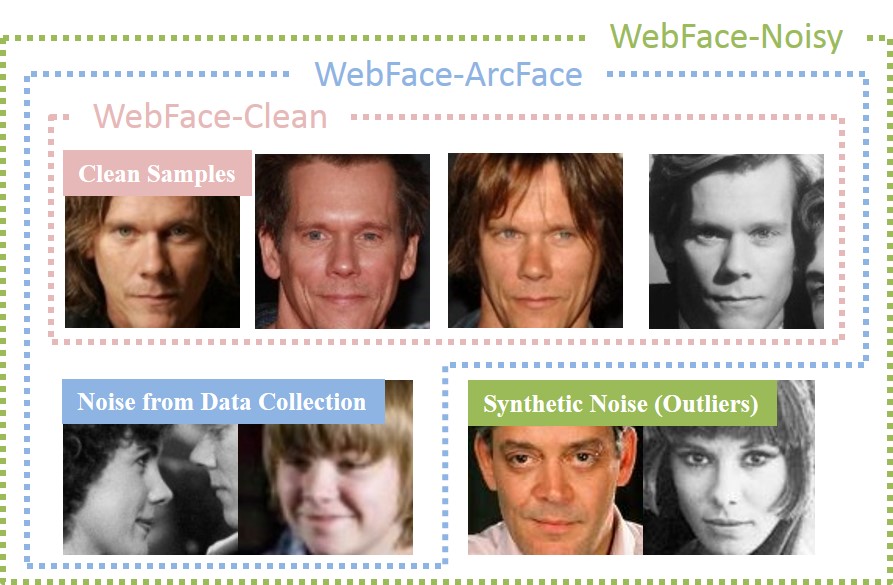}
	\caption{Images of an identity in WebFace-Clean, WebFace-ArcFace and WebFace-Noisy databases. The WebFace-Clean database is a manually cleaned version~\cite{rwebface_wang,deng2019arcface}. The noise in WebFace-ArcFace~\cite{deng2019arcface} database is from the label noise that derive from the collection process of the CASIA-WebFace database~\cite{Yi2014CASIA}. Based on WebFace-ArcFace database, we add images from MS-Celeb-1M database~\cite{guo2016msceleb} evenly across each identity of WebFace-ArcFace database, which means that we incorporate outliers in WebFace-Noisy database.}
	\label{fig:web-noise}
\end{figure}

\begin{table}[htbp]
	\renewcommand\arraystretch{1.05}
	\begin{center}
		\caption{Study on the choice of hype-parameters $a$ and $b$ of SFace (ResNet34). As the noise level increases, parameter $a$ should be larger, \emph{i.e}. ${v_{intra}}\left( {{\theta _{{y_i}}}} \right)$ curves should move to the right, which indicates that the speed of intra-class is decreased more early to prevent overfitting.}
		\label{table:parameters}
		\scalebox{1.0}{
			\begin{tabular}{|c|c|c|c|c|}
				\hline
				\multirow{2}{*}{Noise} & \multicolumn{2}{c|}{Parameters} & \multicolumn{2}{c|}{IJB-C}\\ \cline{2-5} 
				& a & b & FAR=1e-4& FAR=1e-3\\ \hline\hline
				\multirow{4}{*}{\begin{tabular}[c]{@{}c@{}}WebFace-Clean\\ (Noise Level $\approx$ 0\%)\end{tabular}}
				
				&0.81& 1.28& 84.70& 91.71\\ 
				
				& \textbf{0.80} & \textbf{1.28} & 
				\textbf{85.72} & \textbf{92.52} \\ 
				
				& 0.80 & 1.25 & 83.99 & 91.17\\ 
				
				& 0.80 & 1.30 & 85.45  & 92.20\\ \hline
				\multirow{3}{*}{\begin{tabular}[c]{@{}c@{}}WebFace-ArcFace\\ (Noise Level $\approx$ 10\%)\end{tabular}}  
				
				&0.80 & 1.28 & 85.39 & 92.09\\  
				
				& \textbf{0.82} & \textbf{1.28} & \textbf{86.30} & \textbf{92.43} \\ 
				
				& 0.82& 1.25 & 84.17 & 91.32\\ \hline
				
				\multirow{3}{*}{\begin{tabular}[c]{@{}c@{}}WebFace-Noisy\\ (Noise Level $\approx$ 20\%)\end{tabular}}  
				
				&0.82 & 1.28 & 83.97 & 91.36 \\ 
				
				& \textbf{0.84} & \textbf{1.28} & \textbf{84.80} & \textbf{91.84} \\  
				
				& 0.84 & 1.25 & 84.09 & 91.43 \\ \hline
			\end{tabular}
		}
	\end{center}
	
\end{table}

\begin{figure}[htbp]
	\center
	\includegraphics[width=1\linewidth]{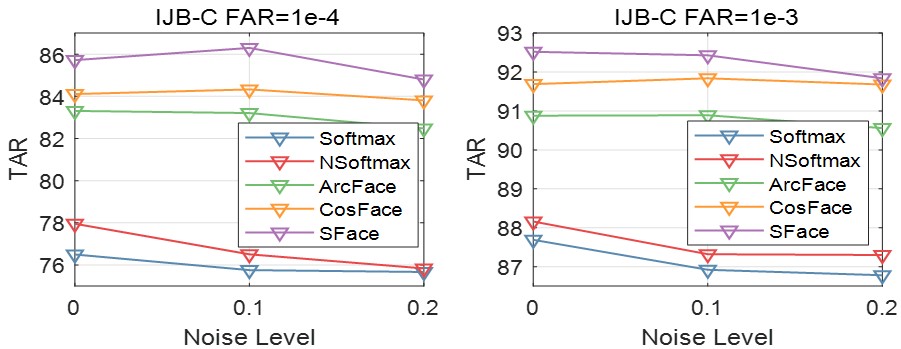}
	\caption{Comparison of verification TAR@FAR=1e-4 and TAR@FAR=1e-3 results on the IJB-C database~\cite{maze2018iarpa} of softmax, NSoftmax, CosFace, ArcFace and SFace models (ResNet34) which are trained with databases of different noise level (WebFace-Clean ($\approx$0\%), WebFace-ArcFace ($\approx$10\%), and WebFace-Noisy ($\approx$20\%)).}
	\label{fig:webface_noise}
\end{figure}

\begin{table}[htbp]
	\renewcommand\arraystretch{1.02}
	\begin{center}
		\caption{Verification performance on LFW~\cite{LFWTech} and YTF~\cite{Wolf2011Face} databases. The state-of-art models in face recognition community are listed for comparison.}
		\label{table:lfw}
		\scalebox{1.0}{
			\begin{tabular}{|c|c|c|c|}
				\hline
				Method&\#Images&LFW&YTF\\
				\hline\hline
				DeepID~\cite{Sun2014Deep}&0.2M&99.47&93.20\\
				DeepFace~\cite{Taigman2014DeepFace}&4.4M&97.35&91.4\\
				VGG Face~\cite{parkhi2015deep}&2.6M&98.95&97.30\\
				FaceNet~\cite{Schroff2015FaceNet}&200M&99.63&95.10\\
				Baidu~\cite{liu2015targeting}&1.3M&99.13&-\\
				Center Loss~\cite{Wen2016A}&0.7M&99.28&94.9\\
				Range Loss~\cite{Zhang2017rangeloss}&5M&99.52&93.70\\
				Marginal Loss~\cite{Deng2017Marginal}&3.8M&99.48&95.98\\
				SphereFace~\cite{Liu2017SphereFace}&0.5M&99.42&95.0\\
				SphereFace+~\cite{liu2018learning}&0.5M&99.47&-\\
				CosFace~\cite{Wang2018CosFace}&5M&99.73&97.6\\
				\hline\hline
				MS1MV2, R100, ArcFace~\cite{deng2019arcface}&5.8M&\bf{99.83}&98.02\\
				MS1MV2, R100, SFace&5.8M&99.82&\bf{98.06}\\
				\hline
			\end{tabular}
		}
	\end{center}
\end{table}

\begin{table}[htbp]
	\renewcommand\arraystretch{1.05}
	\begin{center}
		\caption{Verification performance on on LFW~\cite{LFWTech}, CALFW~\cite{zheng2017CALFW} and CPLFW~\cite{CPLFWTech} databases. The second cell lists results of the open-sourced face recognition models of state-of-art methods. In the third cell, our method is evaluated strictly following ArcFace~\cite{deng2019arcface}.}
		\label{table:cacplfw}
		\scalebox{1.0}{
			\begin{tabular}{|c|c|c|c|}
				\hline
				Method&LFW&CALFW&CPLFW\\
				\hline\hline
				HUMAN-Indivadual&97.27&82.32&81.21\\
				HUMAN-Fusion&99.85&86.50&85.24\\
				\hline\hline
				Center Loss~\cite{Wen2016A}&98.75&85.48&77.48\\
				SphereFace~\cite{Liu2017SphereFace}&99.27&90.30&81.40\\
				VGGFace2~\cite{Cao18}&99.43&90.57&84.00\\
				\hline\hline
				MS1MV2, R100, ArcFace~\cite{deng2019arcface}&99.82&95.45&92.08\\
				MS1MV2, R100, SFace&\bf{99.82}&\bf{96.07}&\bf{93.28}\\
				\hline
			\end{tabular}
		}
	\end{center}
	
\end{table}

\begin{table}[htbp]
	\renewcommand\arraystretch{1.05}
	\begin{center}
		\caption{Face identification and verification evaluation on MegaFace Challenge 1~\cite{kemelmacher2016megaface} using FaceScrub~\cite{Ng2015A} as the probe set. ``Acc.'' refers to the rank-1 face identification accuracy with 1M distractors, and ``Ver.'' refers to the face verification TAR@FAR=1e-6. ``R'' refers to data refinement on both probe set and 1M distractors following~\cite{deng2019arcface}. In the second and third cell, methods are compared in the same setting with ResNet100 models trained on MS1MV2 database~\cite{guo2016msceleb}.}
		\label{table:MF}
		\scalebox{1.0}{
			\begin{tabular}{|c|c|c|c|}
				\hline
				Method&Protocol&Acc.&Ver.\\
				\hline\hline
				FaceNet~\cite{Schroff2015FaceNet}&Large&70.49&86.47\\
				CosFace~\cite{Wang2018CosFace}&Large&82.72&96.65\\
				AdaptiveFace~\cite{liu2019adaptiveface}, R&Large&95.023&95.608\\
				P2SGrad~\cite{zhang2019p2sgrad}, R&Larget&97.25&-\\
				AdaCos~\cite{zhang2019adacos}, R&Large&97.41&-\\
				\hline\hline
				MS1MV2, R100, CosFace~\cite{Wang2018CosFace}&Large&80.56&96.56\\
				MS1MV2, R100, ArcFace~\cite{deng2019arcface}&Large&81.03&96.98\\
				MS1MV2, R100, SFace&Large&\bf{81.15}&\bf{97.11}\\
				\hline\hline
				MS1MV2, R100, CosFace~\cite{Wang2018CosFace}, R&Large&97.91&97.91\\
				MS1MV2, R100, ArcFace~\cite{deng2019arcface}, R&Large&98.35&98.48\\
				MS1MV2, R100, SFace, R&Large&\bf{98.50}&\bf{98.61}\\
				\hline
			\end{tabular}
		}
	\end{center}
	
\end{table}

\begin{table*}[htbp]
	\renewcommand\arraystretch{1.15}
	\begin{center}
		\caption{Face identification and verification evaluation of different methods on the IJB-A~\cite{klare2015pushing} database. In the first cell, experimental results are read from original papers. For comparison, we implement experimental results in the second cell using ArcFace and SFace trained on VGGFace2 and MS-Celeb-1M databases, respectively.}
		\label{table:ijba}
		\scalebox{0.885}{
			\begin{tabular}{|c|c|c|c|c|c|c|c|c|}
				\hline
				\multirow{2}{*}{Method} & \multicolumn{3}{c|}{1:1}       & \multicolumn{5}{c|}{1:N}                     \\ \cline{2-9} 
				& FAR=1e-3 & FAR=1e-2 & FAR=1e-1 & FPIR=0.01 & FPIR=0.1 & Rank-1  & Rank-5  & Rank-10 \\ \hline\hline
				VGGFace~\cite{parkhi2015deep}& 62.00    & 83.40    & 95.40    & 45.40     & 74.80    & 92.50 & 97.20 & 98.30 \\ 
				Template Adaption~\cite{crosswhite2018template}& 83.60    & 93.90    & 97.90    & 77.40     & 88.20    & 92.80 & 97.70 & 98.60 \\
				NAN~\cite{yang2017neural} & 88.10    & 94.10    & 97.80    & 81.70     & 91.70    & 95.80 & 98.00 & 98.60 \\ 
				VGGFace2~\cite{Cao18}& 92.10    & 96.80    & 99.00    & 88.30     & 94.60    & 98.20 & 99.30 & 99.40 \\ 
				FTL~\cite{yin2019feature}& 91.20    & 95.30    & -        & -         & -        & 96.00 & 98.30 & 98.70 \\ 
				UniformFace~\cite{duan2019uniformface}& 92.30    & 96.90    & -        & -         & -        & 97.90 & 98.80 & -     \\ 
				L2-Face~\cite{Ranjan17}& 94.30    & 97.00    & 98.40    & 91.50     & 95.60    & 97.30 & -     & 98.80 \\ 
				Crystal Loss~\cite{ranjan2018crystal}& 94.90    & 96.90    & 98.40    & 91.80     & 95.90    & 97.20 & -     & 98.80 \\ \hline\hline
				VGGFace2, R50, ArcFace~\cite{deng2019arcface}&96.24&98.64&99.51&92.07&97.80&99.19&99.67&99.73\\ 
				VGGFace2, R50, SFace &\bf{96.85}&\bf{98.74}&\bf{99.67}&\bf{92.51}&\bf{98.19}&\bf{99.19}&\bf{99.68}&\bf{99.80}\\  \hline
				
				MS1MV2, R100, ArcFace~\cite{deng2019arcface}& 97.60    & 98.75    & \bf{99.53}    & 93.47     & 98.11    & 98.83 & 99.33 & 99.51 \\ 
				MS1MV2, R100, SFace & \bf{98.02}    & \bf{98.93}    & 99.51    & \bf{94.84}     & \bf{98.50}    & \bf{98.93} & \bf{99.44} & \bf{99.55} \\ \hline
			\end{tabular}
		}
	\end{center}
\end{table*}

\begin{table*}[htbp]
	\renewcommand\arraystretch{1.15}
	\begin{center}
		\caption{Face identification and verification evaluation of different methods on the IJB-C database~\cite{maze2018iarpa}. Experimental results in the first cell are read from original papers, and all the models are trained on VGGFace2 database~\cite{Cao18} or MS-Celeb-1M database~\cite{guo2016msceleb}. For comparison, we implement experimental results in the second cell using ArcFace and SFace trained on VGGFace2 and MS-Celeb-1M databases, respectively.}
		\label{table:ijbc}
		\scalebox{0.885}{
			\begin{tabular}{|c|c|c|c|c|c|c|c|c|c|c|}
				\hline
				\multirow{2}{*}{Method}                                               & \multicolumn{5}{c|}{1:1}       & \multicolumn{5}{c|}{1:N}                         \\ \cline{2-11} 
				& FAR=1e-5 & FAR=1e-4 & FAR=1e-3 & FAR=1e-2 & FAR=1e-1 & FPIR=0.01 & FPIR=0.1 & Rank-1 & Rank-5 & Rank-10 \\ \hline\hline
				VGGFace2, ResNet50~\cite{Cao18}& 73.40    & 82.50    & 90.00  &95.00 & 98.00 & 73.50     & 83.00    & 89.80  & 93.90  & 95.30   \\ 
				VGGFace2, SENet50~\cite{Cao18}& 74.70    & 84.00    & 91.00  &96.00 & 98.70  & 74.60     & 84.20    & 91.20  & 94.90  & 96.20   \\ 
				VGGFace2, MN-v~\cite{xie2018multicolumn}& 75.50    & 85.20    & 92.00  & 96.50 & 98.80  & -         & -        & -      & -      & -       \\ 
				VGGFace2, MN-vc~\cite{xie2018multicolumn}& 77.10    & 86.20    & 92.70 & 96.80 & 98.90   & -         & -        & -      & -      & -       \\ 
				VGGFace2, ResNet50+DCN(Kpts)~\cite{Xie2018ECCV}& -        & 86.70    & 94.00  & 97.90 & 99.70  & -         & -        & -      & -      & -       \\ 
				VGGFace2, ResNet50+DCN(Divs)~\cite{Xie2018ECCV}& -        & 88.00    & 94.40  & 98.10 & 99.80  & -         & -        & -      & -      & -       \\ 
				VGGFace2, SENet50+DCN(Kpts)~\cite{Xie2018ECCV}& -        & 87.40    & 94.40 & 98.10 & 99.80   & -         & -        & -      & -      & -       \\ 
				VGGFace2, SENet50+DCN(Divs)~\cite{Xie2018ECCV} & -        & 88.50    & 94.70  & 98.30 & 99.80   & -         & -        & -      & -      & -       \\ 
				\hline
				MS1M, Inception-ResNet, P2SGrad~\cite{zhang2019p2sgrad}& 87.84    & 92.25    & 95.58  & 97.79 & 99.03  & -         & -        & -      & -      & -       \\ 
				MS1M, Inception-ResNet, AdaCos~\cite{zhang2019adacos}& 88.03    & 92.40    & 95.65  & 97.72 &  99.06 & -         & -        & -      & -      & -       \\ \hline\hline
				VGGFace2, R50, ArcFace~\cite{deng2019arcface}&86.03&92.12&95.93 &98.23&\bf{99.34}&79.50&89.53&94.75&96.94&97.64\\ 
				VGGFace2, R50, SFace  
				&\bf{87.08}&\bf{93.12}&\bf{96.50} &\bf{98.34} & 99.25 &\bf{82.84}&\bf{90.69}&\bf{95.01}&\bf{96.97}&97.55\\ \hline			
				
				MS1MV2, R100, ArcFace~\cite{deng2019arcface}& 93.15    & 95.65    & 97.20  & 98.18 & \bf{99.01}  & 90.32     & 94.52    & 95.72  & 97.10  & 97.47   \\ 
				MS1MV2, R100, SFace & \bf{94.21}    & \bf{96.11}    & \bf{97.50}  & \bf{98.33} & 99.00  & \bf{92.41}     & \bf{95.17}    & \bf{96.21}  & \bf{97.41}  & \bf{97.76}   \\ \hline
			\end{tabular}
		}
	\end{center}
\end{table*}

\begin{figure*}[htbp]
	\center
	\includegraphics[width=1\linewidth]{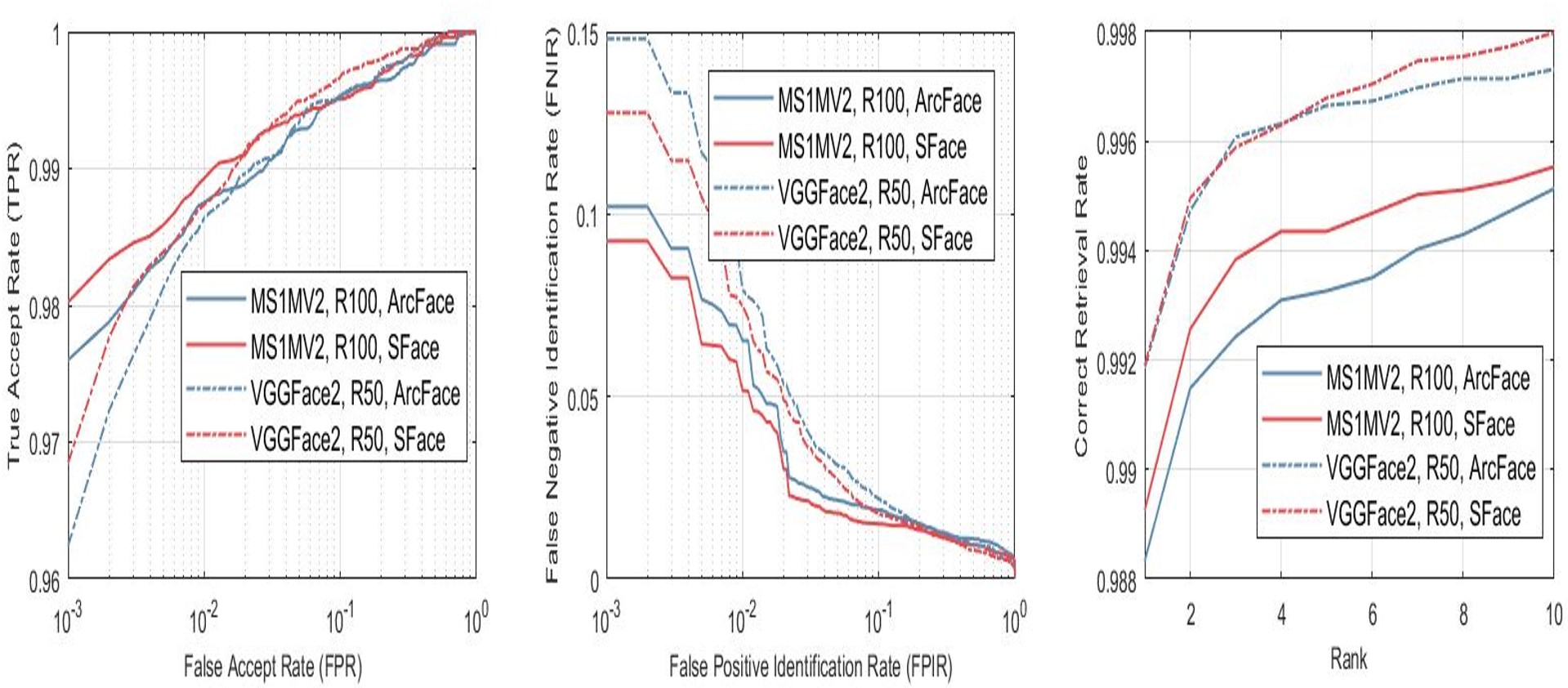}
	\caption{Comparison of ArcFace and SFace models on the IJB-A  database~\cite{klare2015pushing}. Left: ROC (higher is better). Middle: DET (lower is better). Right: CMC (higher is better). Our method is represented using red color.}
	\label{fig:pic_ijba}
\end{figure*}

\begin{figure*}[htbp]
	\center
	\includegraphics[width=1\linewidth]{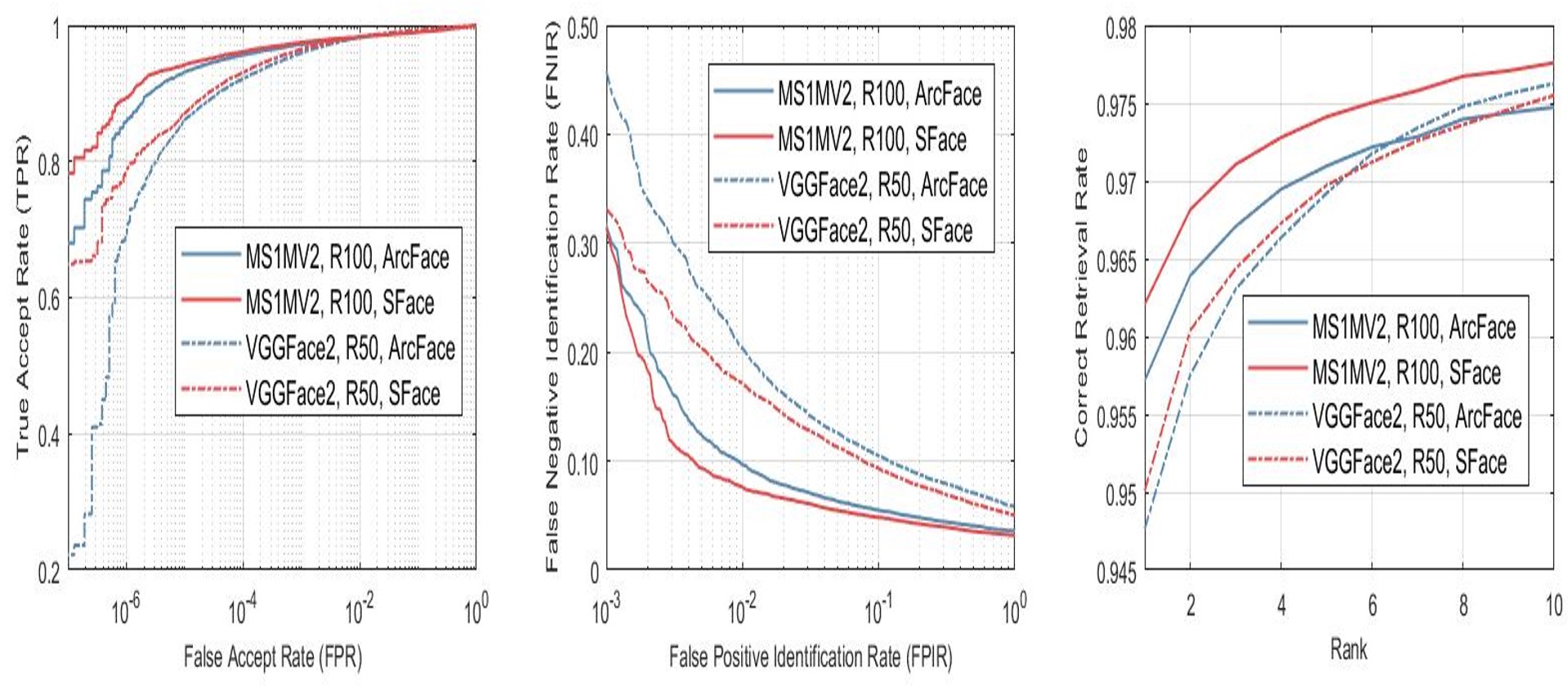}
	\caption{Comparison of ArcFace and SFace models on the IJB-C database~\cite{maze2018iarpa}. Left: ROC (higher is better). Middle: DET (lower is better). Right: CMC (higher is better). Our method is represented using red color.}
	\label{fig:pic_ijbc}
\end{figure*}

\subsubsection{Experiment on the Noise-Controlled WebFace Database}
To further evaluate our method on the training databases with low level noise, we train deep face models under noise-controlled settings. Specifically, we use three databases with different noise level. (1) Since we have the manual refined image list released by~\cite{rwebface_wang}, we first clean the ArcFace version~\cite{deng2019arcface} of CASIA-WebFace database. Finally, we obtain a manually cleaned version of CASIA-WebFace database (0.45M images from 10,572 identities). This database is named as WebFace-Clean. (2) Then, ArcFace version~\cite{deng2019arcface} of CASIA-WebFace database (0.49M images from 10,572 identities) is used as first noise level database. We name this database as WebFace-ArcFace. (3) Finally, we augment the ArcFace version~\cite{deng2019arcface} of CASIA-WebFace database with synthesis images. We add images from MS-Celeb-1M database~\cite{guo2016msceleb} evenly across each identity of WebFace-ArcFace database. That is to say, we incorporate outliers in this training database. The database is referred to as WebFace-Noisy. We use this setting because in practice, outliers noise is a more common type of label noise than label flip noise. The noise level of WebFace-Clean, WebFace-ArcFace and WebFace-Noisy is approximately 0\%, 10\% and 20\%, respectively. Some identities of WebFace-ArcFace and WebFace-Noisy databases are shown in Figure~\ref{fig:web-noise}. Note that the 10\% noise in WebFace-ArcFace database is from the label noise that derive from the collection process of the CASIA-WebFace database. While the 20\% label noise in WebFace-Noisy contains 10\% noise in WebFace-ArcFace and other 10\% synthetic outliers. 

We train ResNet34 models on WebFace-Clean, WebFace-ArcFace and WebFace-Noisy databases supervised by softmax, NSoftmax, CosFace, ArcFace and SFace. The experimental results are shown in Figure~\ref{fig:webface_noise}, which demonstrates the robustness of SFace to low level label noise. We also list the choice of hyper-parameters of SFace in Table~\ref{table:parameters}. We can conclude that parameter $a$ should be larger, \emph{i.e}. ${v_{intra}}\left( {{\theta _{{y_i}}}} \right)$ curves should move to the right, as the noise level increases, which indicates that the speed of intra-class is decreased more early to prevent overfitting. Although inter-class parameters $b$ are also important for training, we find the optimal groups of them are the same for the three training databases, the reason may be that noisy data is relatively balanced across all identities. Another interesting phenomenon is that the model have similar performance on WebFace-Clean and WebFace-ArcFace. This result indicates that the manual cleaned data by human annotations actually has limited influence on these face models.

\subsection{Evaluation Results on Several Benchmarks}
We first evaluate our method on LFW~\cite{LFWTech} and YTF~\cite{Wolf2011Face}. For fair comparison, we train models using ResNet100 on MS1MV2 database~\cite{guo2016msceleb}, strictly following the settings in~\cite{deng2019arcface}. MS1MV2 database is a refined version of MS-Celeb-1M database~\cite{guo2016msceleb}, cleaned by insightface~\cite{deng2019arcface}. MS1MV2 database contains 5.8M images of 85,742 celebrities. We use this semi-artificial cleaned face database as a large-scale training database to further evaluate our method. For SFace trained on MS1MV2 database, the hyper-parameters $a$, $b$ are set as 0.90 and 1.20. The experimental results on LFW and YTF are shown in Table~\ref{table:lfw}. SFace model trained on MS1MV2 database with ResNet100 obtains comparable results as the baseline method such as CosFace~\cite{Wang2018CosFace} and ArcFace~\cite{deng2019arcface}. We report the performance on CALFW~\cite{zheng2017CALFW} and CPLFW~\cite{CPLFWTech} databases in Table~\ref{table:cacplfw}. As shown in Table~\ref{table:cacplfw}, SFace outperforms both human performance and the advanced deep face models on CALFW and CPLFW databases by a significant margin. 

Then, we evaluate our method on MegaFace database~\cite{kemelmacher2016megaface} including both the original MegaFace database and the refined version~\cite{deng2019arcface}. We report the rank-1 face identification accuracy with 1M distractors, and the face verification TAR@FAR=1e-6, shown in Table~\ref{table:MF}. In the second and third cell, methods are compared in the same setting with ResNet100 models trained on MS1MV2 database. As reported in Table~\ref{table:MF}, our method shows superiority over CosFace and ArcFace on both identification and verification settings on MegaFace challenge. 

Finally, we evaluate our method on IJB-A~\cite{klare2015pushing} and IJB-C~\cite{maze2018iarpa} databases on both identification and verification settings. Our method is compared with ArcFace using the same databases and models, other results are cited from the original papers. For fair comparison, we also train ResNet50 models on VGGFace2 database~\cite{Cao18} following~\cite{deng2019arcface}. VGGFace2 training database has 3.13 million images of 8,631 identities, and has large variations in pose, age, illumination, ethnicity and profession. For SFace model trained on VGGFace2 database, the hyper-parameters $a$, $b$ are set as 0.88 and 1.25. The results on IJB-A database are exhibited in Table~\ref{table:ijba} and Figure~\ref{fig:pic_ijba}. The results on IJB-C database are shown in Table~\ref{table:ijbc} and Figure~\ref{fig:pic_ijbc}. For verification, we report TAR@FAR (ROC curves, higher is better). For identification, the performance is reported using TPIR@FPIR (DET curve, lower is better) and Rank-N accuracy (CMC curve, higher is better). Compared with ArcFace models trained on both VGGFace2 and MS1MV2 databases, our method performs better in both identification and verification settings, especially the TAR at very low FAR, which demonstrates the effectiveness and superiority of SFace. 

\section{Conclusion}
\label{sec:conclusion}
In this paper, different from previous works which minimize the intra-class distances and maximize the inter-class distance, we introduce a new idea which aims to optimize intra-class and inter-class distance to some extent for the purpose of mitigating overfitting problems to the imperfect training databases. To carry out this idea, we propose a new loss function SFace to improve the performance of models in the robust unconstrained face recognition. SFace imposes intra-class and inter-class constraints on a hypersphere manifold with precisely controlled intra-class and inter-class gradients so that intra-class and inter-class distances are optimized to some extent. To promote further understanding of SFace, we explain the relationship to softmax based loss functions, and show that, compared with softmax based loss, the advantage of SFace is the precisely control ability of both intra-class and inter-class optimization. The proposed SFace makes a better balance between underfitting and overfitting, and further improves the generalization ability of deep face models. Experiments on several benchmarks including LFW, YTF, CALFW, CPLFW, MegaFace, IJB-A and IJB-C databases, have demonstrated the effectiveness and superiority of our method. 

\section*{Acknowledgment}
This work was supported by Canon
Information Technology (Beijing) Co., Ltd. under Grant
No. OLA19023, and supported by BUPT Excellent Ph.D. Students Foundation CX2020201.

\ifCLASSOPTIONcaptionsoff
\newpage
\fi
	
\bibliographystyle{IEEEtran}
\bibliography{egbib}
	
\end{document}